\documentclass[letterpaper, 10 pt, journal, twoside]{ieeetran}
\UseRawInputEncoding
\usepackage[ruled,vlined,linesnumbered]{algorithm2e}
\usepackage{amsmath,amsfonts}
\usepackage{caption}
\usepackage{booktabs}
\usepackage{multirow}
\usepackage{makecell}
\usepackage{diagbox}
\usepackage{array}
\usepackage{booktabs}
\usepackage{bigstrut}
\usepackage{color}
\usepackage{xcolor}
\definecolor{BLACK}{rgb}{0,0,0}
\usepackage{colortbl}
\usepackage[colorlinks,
            linkcolor=green,
            anchorcolor=green,
            citecolor=green]{hyperref}
            
\usepackage{authblk}
\usepackage{array}
\usepackage[caption=false,font=normalsize,labelfont=sf,textfont=sf]{subfig}
\usepackage{textcomp}
\usepackage{stfloats}
\usepackage{url}
\usepackage{verbatim}
\usepackage{graphicx}
\usepackage{cite}
\usepackage{bbm}
\definecolor{mygreen}{rgb}{.75,1,.75}
\usepackage{etoolbox}
\makeatletter
\patchcmd{\@makecaption}
{\scshape}
{}
{}
{}
\newcommand{\customref}[2]{\hyperref[#2]{#1}}
\usepackage{pifont}

\usepackage[capitalize]{cleveref}
\makeatother

\title{
	Fast-Poly: A Fast Polyhedral \textcolor{black}{Framework} For 3D Multi-Object Tracking
}

\author{Xiaoyu Li$^\dagger$, Dedong Liu$^\dagger$, \textcolor{black}{Yitao Wu$^\dagger$, Xian Wu$^\dagger$, Lijun Zhao$^{*}$,} Jinghan Gao% <-this % stops a space
	\thanks{$\dagger$: These authors contributed equally to this work.                 *: Corresponding author.} % <-this % stops a space
        \thanks{All authors are with State Key Laboratory of Robotics and System, Harbin Institute of Technology, Harbin 150006, China.}
        }% <-this % stops a space
\begin{document}
	\markboth{}
	{Li \MakeLowercase{\textit{et al.}}: Fast-Poly: A Fast Polyhedral \textcolor{black}{Framework} For 3D Multi-Object Tracking} 
	\maketitle

	%%%%%%%%%%%%%%%%%%%%%%%%%%%%%%%%%%%%%%%%%%%%%%%%%%%%%%%%%%%%%%%%%%%%%%%%%%%%%%%%
	\begin{abstract}
        3D Multi-Object Tracking (MOT) captures stable and comprehensive motion states of surrounding obstacles, essential for robotic perception. 
        However, current 3D trackers face issues with accuracy and latency consistency.
        In this paper, we propose Fast-Poly, a fast and effective filter-based method for 3D MOT.
        Building upon our previous work Poly-MOT, Fast-Poly addresses object rotational anisotropy in 3D space, enhances local computation densification, and leverages parallelization technique, improving inference speed and precision.
        Fast-Poly is extensively tested on two large-scale tracking benchmarks with Python implementation.
        On the nuScenes dataset, Fast-Poly achieves new state-of-the-art performance with 75.8\% AMOTA among all methods and can run at 34.2 FPS on a personal CPU.
        On the Waymo dataset, Fast-Poly exhibits competitive accuracy with 63.6\% MOTA and impressive inference speed (35.5 FPS).
        The source code is publicly available at \href{https://github.com/lixiaoyu2000/FastPoly}{https://github.com/lixiaoyu2000/FastPoly}.

	\end{abstract}
 
	%%%%%%%%%%%%%%%%%%%%%%%%%%%%%%%%%%%%%%%%%%%%%%%%%%%%%%%%%%%%%%%%%%%%%%%%%%%%%%%%
 
        \section{INTRODUCTION}
        \IEEEPARstart{3D}{Multi-Object Tracking} filters and recalls obstacle observations, being a crucial component in autonomous driving and robot perception systems.
        Recent advances in 3D MOT can be attributed to the success of the Tracking-By-Detection (TBD) framework\footnote{Top 5 methods on \href{https://www.nuscenes.org/tracking?externalData=all&mapData=all&modalities=Any}{nuScenes} are all adhere to the TBD framework.}.
        Despite the advancements, these filter-based works~\cite{li2023poly, li2023camo, kim2021eagermot, pang2022simpletrack, gwak2022minkowski, sadjadpour2023shasta, zaech2022learnable} currently suffer from the consistency of accuracy and latency on large-scale datasets (nuScenes~\cite{caesar2020nuscenes}, Waymo~\cite{sun2020scalability}, etc.), as shown in Fig. \ref{fig:f0}.
        Specifically, these datasets mainly bring the proliferation of tracking agents, magnifying four primary shortcomings of filter-based methods:
        
        \begin{itemize}
        
        \item \textbf{Object is rotated in 3D space.} 
        Geometry-based metrics (IoU~\cite{weng20203d, wang2022deepfusionmot}, GIoU~\cite{li2023camo, li2023poly, pang2022simpletrack}) provide high interpretability and accurately capture inter-object affinity, are widely utilized.
        However, according to statistics in Table \ref{table:nu_metric_time}, 3D object rotation leads to time-consuming polygon intersection and convex hull solution steps in these metrics. 
        Some works~\cite{kim2021eagermot, PC3T} employ custom Euclidean distance to reduce overhead, but accuracy is lost due to the lack of spatial relationship considerations. 
        
        \item \textbf{The tracking pipeline is from a global perspective.}
        Modeling global object similarity for pre-processing and association is a common practice in the field~\cite{li2023camo, pang2022simpletrack, weng20203d, PC3T, wang2022deepfusionmot, li2023poly, kim2021eagermot}.
        However, inter-observation redundancy and trajectory-observation correlation are only feasible in a limited local space, leading to numerous invalid computations. 
        \cite{li2023poly, li2023camo, pang2022simpletrack} emphasize restricting false positive (FP) agents from influencing cost computations.
        \cite{li2023poly} also applies multiple criteria to avoid distinct categories matching.
        Despite these efforts, the fact that \textit{distant objects are unrelated} is widely ignored.

        \begin{figure}[t]
        \centering
        %\framebox{\parbox{3in}{We suggest that you use a text box to insert a graphic (which is ideally a 300 dpi TIFF or EPS file, with all fonts embedded) because, in a document, this method is somewhat more stable than directly inserting a picture.}}
        \includegraphics[width=\linewidth]{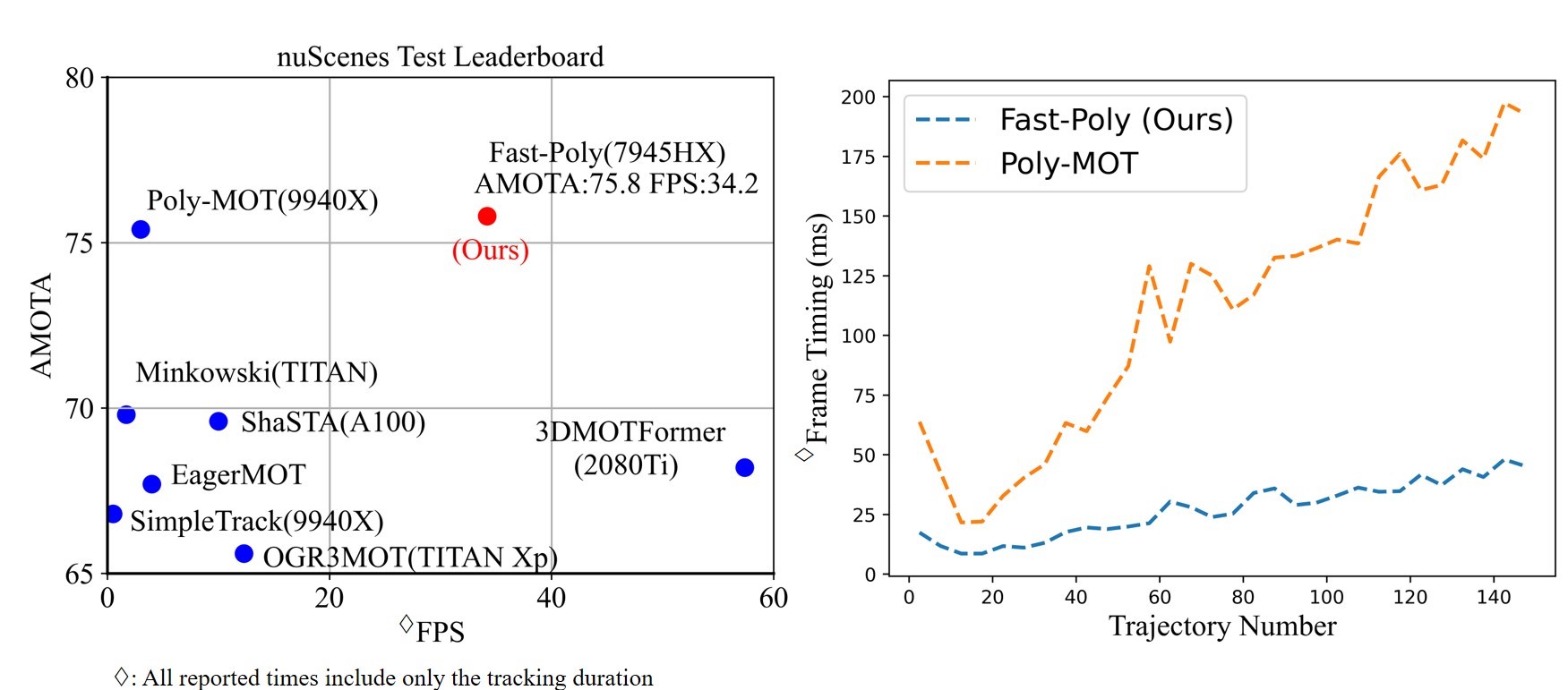}
        \caption{\textbf{Left}: The comparison of the accuracy and latency between our method and advanced trackers on \href{https://www.nuscenes.org/tracking?externalData=all&mapData=all&modalities=Any}{nuScenes test leaderboard}.
        The closer to the top right, the better the performance.
        Fast-Poly also exhibits superior performance on the \href{https://waymo.com/open/challenges/2020/3d-tracking/}{Waymo test leaderboard} with 63.6\% MOTA and 35.5 FPS.
        \textbf{Right}: The time consumption curve of average frame timing with the objects increasing between our proposed Fast-Poly and baseline method Poly-MOT on nuScenes val set.}
        \label{fig:f0}
        \end{figure}
        
        \item \textbf{Matrix calculations in filters are heavy.}
        Trajectories typically represented as high-dimensional vectors~\cite{li2023camo, pang2022simpletrack, weng20203d, PC3T, wang2022deepfusionmot, li2023poly, kim2021eagermot}.
        Nevertheless, integrating time-invariant states in the Kalman Filter (KF) leads to redundant computations involving zeros.
        \cite{liu2023fasttrack} accelerates 2D position estimation through a priori split calculation and CUDA operator.
        However, in 3D MOT, the abundance of states and non-linear motion models~\cite{li2023poly} make a similar matrix-to-quadratic form transformation challenging.
        Furthermore, rigid count-based lifecycle management hinders real-time performance due to matrix operations applied to all active tracklets, including the coasted FP tracklets.
        
        \item \textbf{The TBD framework is serial logic inherently.}
        Existing methods rely on a module-by-module program execution order due to the extensive fusion calculations between observations and tracklets within the TBD framework. 
        This conventional view overlooks the potential for localized parallel processing, hindering efficiency.
        
        \end{itemize}
        
        Towards these issues, we propose Fast-Poly, a fast and effective polyhedral framework for 3D MOT.
        % handle rotation (alignment)
        Specifically, to eliminate the object rotation impact, we introduce an elegant solution: Align rotated objects and then employ 2D-like parallel IoU operations for accelerated similarity calculations.
        % handle global perspective
        To avoid invalid global affinity computations, we introduce the voxel mask to calculate valuable costs quickly based on coarse-to-fine criteria.
        % handle numerous (FP) and heavy (high-dim state) matrix ops (Densification)
        To minimize the KF burden, we first incorporate a lightweight filter to decouple and manage time-invariant states, streamlining the trajectory vectors.
        Additionally, we propose a confidence-count mixed lifecycle strategy to flexibly terminate the trajectory, mitigating FP tracking agents in the maintenance list.
        % handle serial bottleneck (Parallelization)
        As a practical solution, we parallelize pre-processing and motion prediction based on their independent relationships, alleviating the serial bottleneck.
        
        Fast-Poly is learning-free and can perform fast, accurate tracking with limited resources (only CPU).
        Fast-Poly builds upon our previous work Poly-MOT~\cite{li2023poly}, an advanced tracker optimized for multi-category scenes.
        With the Python implementation, Fast-Poly is fully assessed on two autonomous driving datasets (nuScenes~\cite{caesar2020nuscenes}, Waymo~\cite{sun2020scalability}).
        \textbf{On nuScenes, Fast-Poly establishes a new state-of-the-art with 75.8\% AMOTA and 34.2 FPS among all methods}, accompanied by a 5$\times$ faster inference speed than baseline.
        On Waymo, Fast-Poly attains competitive performance with 63.6\% MOTA and impressive speed (35.5 FPS).
        Our main contributions include:
        
        \begin{itemize}
        \item We propose Fast-Poly, a filter-based 3D MOT method that combines high real-time performance with outperform accuracy across two large-scale autonomous driving datasets (nuScenes~\cite{caesar2020nuscenes}, Waymo~\cite{sun2020scalability}).
        \item We leverage rotated object alignment, local computation densification, and module parallelization techniques to solve the real-time dilemma of filter-based methods while improving accuracy.
        \item We achieve state-of-the-art tracking performance on the nuScenes test leaderboard among all methods with 75.8\% AMOTA and 34.2 FPS.
        \item Our code is made publicly available, aiming to serve as a strong baseline for the community.
        \end{itemize}

	\section{Related Work}
	
        \subsection{3D Multi-Object Tracking}        
        Current 3D MOT~\cite{benbarka2021score, li2023poly, li2023camo, ding20233dmotformer, zhang2023bytetrackv2, gwak2022minkowski, kim2021eagermot, pang2022simpletrack, PC3T, wang2022deepfusionmot, weng20203d, sadjadpour2023shasta, mutr3d, pftrack, yin2021center, zaech2022learnable} benefits from advancements in 2D MOT~\cite{liu2023fasttrack, zhou2020tracking, bernardin2006multiple, luiten2021hota} and 3D detection~\cite{huang2021bevdet, liu2023bevfusion, chen2022scaling, yin2021center, wu2022casa, shi2020pv}.
        AB3DMOT~\cite{weng20203d} pioneers the expansion of the TBD framework into 3D space, introducing a simple yet effective baseline.
        Poly-MOT~\cite{li2023poly} proposes category-specific ideas to introduce a strong tracker based on multiple non-linear models and similarity metrics, achieving SOTA performance on nuScenes.
        With Neural Network (NN) and a post-processing manner, some methods~\cite{gwak2022minkowski, sadjadpour2023shasta, ding20233dmotformer, zaech2022learnable} utilize the features output by the pre-trained backbone~\cite{yin2021center} to regress the affinity or even the tracking confidence~\cite{sadjadpour2023shasta}.
        Leveraging the occlusion robustness of LiDAR and the far-sightedness of cameras, EagerMOT~\cite{kim2021eagermot} performs a detection-level fusion to combine 2D and 3D detectors.
        CAMO-MOT~\cite{li2023camo} introduces an occlusion head to classify the visual occlusion status and implements a confidence-involved LiDAR association.
        % JMOT~\cite{huang2021joint} and mmMOT~\cite{mmMOT_ICCV} develop a feature-level fusion multi-modal tracker, generating detections and association based on the Joint Detection and Tracking (JDT) framework.
        Being economical and an end-to-end pipeline, multi-camera trackers~\cite{mutr3d, pftrack} receive attention and applications.
        Following the Tracking-By-Attention (TBA) framework, these trackers implement spatial-temporal modeling based on self/cross-attention and implicitly learn trajectory motion in a self-refinement paradigm.
        The trade-off between computational overhead, real-time performance, and tracking accuracy is the primary research direction in the current landscape.

        \subsection{Tracking-By-Detection}

        \textbf{Core Idea.} 
        Taking advanced detection results as input, tracking is handled independently on the tracker-end entirely in a post-processing manner.
        The TBD framework features a well-established pipeline, dividing 3D MOT into four components: pre-processing, estimation, association, and lifecycle.
        
        \textbf{Pre-processing.}
        This module recalls high-quality observations from raw detections based on spatial information between detections (Non-Maximum Suppression~\cite{pang2022simpletrack, li2023poly, li2023camo}, NMS) or confidence (Score Filter~\cite{li2023poly, li2023camo}, SF).
        However, the global perspective inherent in the process leads to redundant computations. 
        In contrast, our method employs the voxel mask, effectively bypassing these calculations.
        
        \textbf{Estimation.}
        This module utilizes filters (Linear Kalman Filter~\cite{kim2021eagermot, li2023camo, pang2022simpletrack, PC3T, wang2022deepfusionmot, weng20203d}, Extended Kalman Filter~\cite{li2023poly}, Point Filter~\cite{benbarka2021score, yin2021center}, etc.), along with motion models, to perform two functions:
        (1) Predict the alive tracklets to achieve temporal alignment and association with the detection.
        (2) Update the matched tracklets with the corresponding observations, preparing prior information for downstream.
        Nevertheless, the real-time challenge emerges when managing numerous tracklets, as the filter involves heavy matrix operations on each tracklet.
        Therefore, we decouple and filter time-invariant states using a lightweight filter to alleviate the computational burden.

        \textbf{Association.}
        As the core of the system, this module establishes tracklet-observation similarity and resolves matching correspondence.
        Geometry-based (IoU~\cite{weng20203d, wang2022deepfusionmot}, GIoU~\cite{li2023camo, li2023poly, pang2022simpletrack}, Euclidean~\cite{benbarka2021score, PC3T, kim2021eagermot}, NN distance~\cite{ding20233dmotformer, sadjadpour2023shasta, zaech2022learnable, gwak2022minkowski}, etc.) and appearance-based are the commonly used affinity metrics.
        The former modeling process, reliant solely on spatial information, is mostly unsupervised and robust to occlusion.
        The latter, robust to inaccurate depth, can describe distant objects.
        Under the one-to-one assumption, the optimal assignment~\cite{kuhn1955hungarian} algorithm is then used to obtain matching pairs.
        However, rotation in 3D space impairs the parallelization of IoU-like metrics, introducing severe latency.
        Fast-Poly proposes to leverage alignment to solve this dilemma.

        \textbf{Lifecycle.}
        This module initializes, terminates, and merges tracklets based on the count-based strategy~\cite{kim2021eagermot, li2023camo, li2023poly, pang2022simpletrack, wang2022deepfusionmot, weng20203d} or the confidence-based~\cite{benbarka2021score, li2023poly, pang2022simpletrack, PC3T} strategy.
        The former has simple logic and strong robustness.
        The latter is more flexible in processing based on heuristic functions.
        In response to the inherent rigidity issue of the former approach, we employ score refinement to flexible terminate trajectories. 
        In addition, we leverage trajectory average scores and a novel prediction to bolster occlusion robustness and enhance the estimation accuracy of the confidence-based strategy.

        % \end{itemize}
        \begin{figure*}[t]
        \vspace{0.5em}
        \centering
        %\framebox{\parbox{3in}{We suggest that you use a text box to insert a graphic (which is ideally a 300 dpi TIFF or EPS file, with all fonts embedded) because, in a document, this method is somewhat more stable than directly inserting a picture.}}
        \includegraphics[width=1\textwidth]{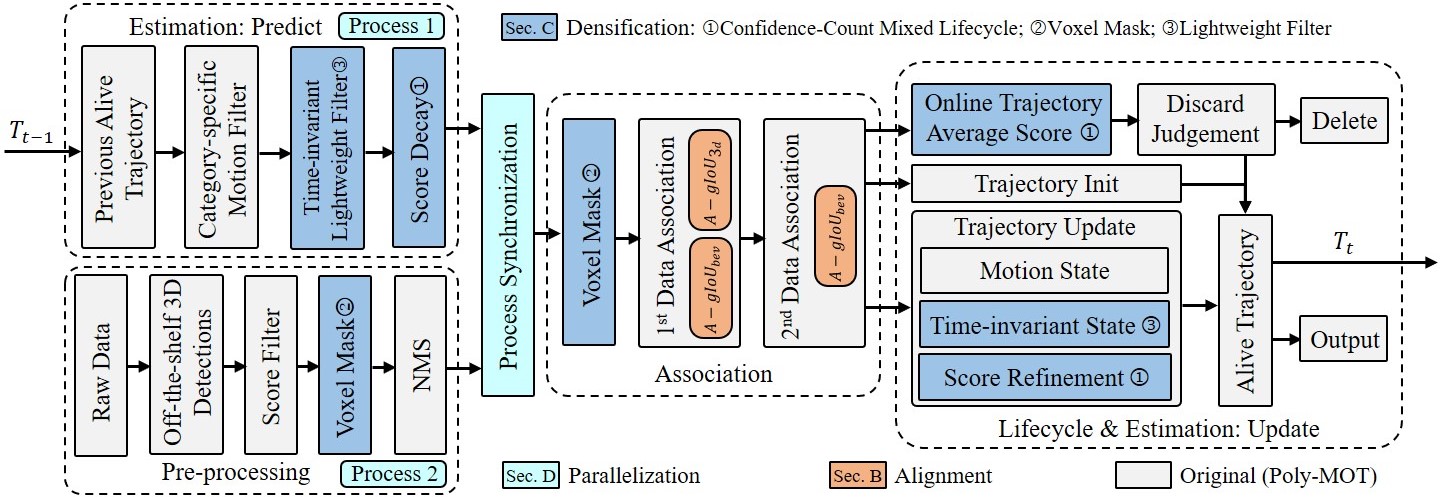}
        \caption{The pipeline of our proposed method.
        Our structure design is illustrated in \cref{Architecture}.
        Real-time improvements to the baseline~\cite{li2023poly} are highlighted in distinct colors.
        \textbf{\textcolor[RGB]{244, 177, 131}{Orange}} denotes \textcolor{black}{the \textit{Alignment} to reduce the computational complexity of affinity calculations for rotated objects.}
        \textbf{\textcolor[RGB]{157, 195, 230}{Blue}} denotes the \textit{Densification} to increase computational efficiency.
        % \textcolor{blue}{In the \textit{Densification},
        % \normalsize{\textcircled{\footnotesize{1}}} denotes Confidence-Count Mixed Lifecycle,
        % \normalsize{\textcircled{\footnotesize{2}}} denotes Voxel Mask,
        % \normalsize{\textcircled{\footnotesize{3}}} denotes Lightweight Filter.}
        \textbf{\textcolor[RGB]{204, 255, 255}{Cyan}} denotes \textcolor{black}{the \textit{Parallelization} to execute pre-processing and motion prediction simultaneously, enhancing computational efficiency.}
        }
        \label{fig:f1}
        \vspace{-1.5em}
        \end{figure*}

	\section{Fast-Poly}

        \subsection{Overall Architecture}
        \label{Architecture}

        Despite sharing a similar pipeline, Fig. \ref{fig:f1} illustrates the distinctions between our proposed method and Poly-MOT~\cite{li2023poly}.
        % pre-processing and predict
        At each frame $t$, Fast-Poly employs two independent computing processes (\cref{parallel}) to filter the 3D detections $D_{t}$ and predict existing trajectories $T_{t, t-1}$.
        Specifically, SF and NMS filters are leveraged to process the detections.
        Besides, the motion (time-variable), score (\textcolor{red}{\customref{Section III-C1}{cb-life}}), and time-invariant (\customref{Section III-C3}{lwfilter}) states of trajectories are predicted by corresponding filters.
        % association and matching
        $D_{t}$ and $T_{t, t-1}$ are then subjected to construct cost matrices for two-stage association.
        Notably, during NMS and matching, our proposed voxel mask (\customref{Section III-C2}{voxel_mask}) and a novel geometry-based metric (\cref{agiou}) accelerate computing.
        Hungarian algorithm~\cite{kuhn1955hungarian} is then employed to obtain matched pairs $DT_{t}$, unmatched detections $D_{t}^{um}$, unmatched tracklets $T_{t-1}^{um}$.
        % update state
        To relieve the matrix dimension, time-invariant and time-variable states in matched tracklets $T_{t-1}^{m}$ are updated with corresponding observations $D_{t}^{m}$ based on our proposed lightweight filter (\customref{Section III-C3}{lwfilter}) and Extended Kalman Filter (EKF), respectively.
        $T_{t-1}^{m}$ scores are also refined leveraging the confidence-count mixed lifecycle (\customref{Section III-C1}{cb-life}).
        % init and dead
        $D_{t}^{um}$ are then initialized as newly active tracklets.
        To flexibly determine FP agents in $T_{t-1}^{um}$, we soft-terminate the mismatch tracklets by considering their max-age and online average refined scores.
        % next frame
        The remaining tracklets $T_{t}$ are sent to downstream tasks and implemented for next frame tracking.
        \vspace{-0.5em}

        \begin{figure*}[t]
        \centering
        %\framebox{\parbox{3in}{We suggest that you use a text box to insert a graphic (which is ideally a 300 dpi TIFF or EPS file, with all fonts embedded) because, in a document, this method is somewhat more stable than directly inserting a picture.}}
        \includegraphics[width=\textwidth]{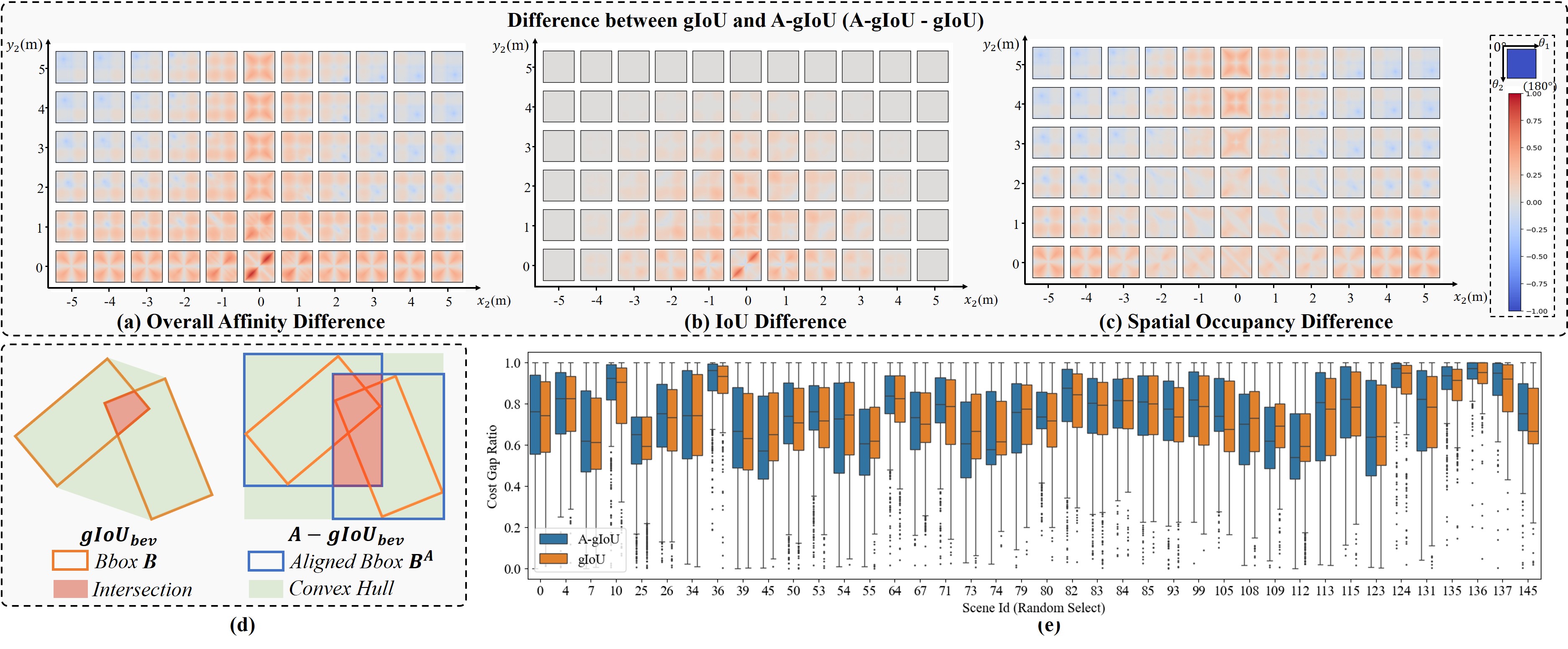}
        \caption{\textcolor{black}{
        \textbf{Top}: Affinity difference between $A\text{-}gIoU$ and $gIoU$ for identical bbox pairs under varying geometric conditions in the 3D space.
        \textbf{(a)}: overall affinity difference.
        \textbf{(b)}: $IoU$ difference.
        \textbf{(c)}: spatial occupancy difference.
        \textbf{Bottom-Left}: The calculation process of distinct metrics in the BEV space.
        \textbf{Bottom-Right}: The contributions of CGR across all trajectories within randomly sampled scenes from nuScenes of $A\text{-}gIoU$ and $gIoU$, employing our tracker with CenterPoint~\cite{yin2021center} detector.}}
        \label{fig:f2}
        \vspace{-1.5em}
        \end{figure*}

	\subsection{Alignment}
        \label{Alignment}
        This section details how Fast-Poly mitigates high latency caused by object rotations through alignment, covering implementation and principles.
        
        \textbf{GIoU.}
        % Rotating angle mainly brings a huge delay to the inter-object affinity computation, since geometry-based metrics dominate advanced filter-based methods~\cite{li2023poly, pang2022simpletrack, li2023camo, weng20203d, wang2022deepfusionmot}.
        Without loss of generality, we optimize the widely employed metric in nuScenes SOTA trackers~\cite{li2023poly, li2023camo} Generalized Intersection over Union $gIoU$, which is formulated as:
        \begin{equation}
        \label{eq:giou}
            gIoU(B_{i}, B_{j}) = \frac{\Lambda(B_{i} \cap B_{j})}{\Lambda(B_{i} \cup B_{j})} + \frac{\Lambda(B_{i} \cup B_{j})}{\Lambda_{Hull}(B_{i}, B_{j})} - 1, 
        \end{equation}
        where $B$ is the 3D bounding box representing the detection or tracklet.
        According to the representation space, $\Lambda(\cdot)$ is the area function in the BEV space and the volume function in the 3D space.
        $B_{i} \cap B_{j}$, $B_{i} \cup B_{j}$ and $\Lambda_{Hull}$ are the intersection, union and convex hull computed by $B_{i}$ and $B_{j}$, respectively.
        
        \textbf{A-GIoU.}
        Varied from the pixel plane, the rotation causes overlap and convex hull irregular polygons instead of axis-aligned rectangles.
        The complexity of the well-known Sutherland-Hodgman~\cite{sutherland1974reentrant} and Graham scan~\cite{graham1972efficient} algorithm are $\mathcal{O}(n^{2})$ and $\mathcal{O}(n\log{n})$, where $n$ is the number of input points.
        In addition, the sum calculation time increases rapidly with the growth of detections and tracklets.
        On the contrary, the solution complexity of the axis-aligned bounding box is only \textcolor{black}{ $\mathcal{O}(n)$}.
        To this end, we proposed the Aligned Generalized Intersection over Union ($A\text{-}gIoU$), which is formulated as:
        \label{agiou}
        \begin{equation}
        \label{eq:agiou}
            A\text{-}gIoU(B_{i}, B_{j}) = \frac{\Lambda(B^{A}_{i} \cap B^{A}_{j})}{\Lambda(B^{A}_{i} \cup B^{A}_{j})} + \frac{\Lambda(B^{A}_{i} \cup B^{A}_{j})}{\Lambda_{Hull}(B^{A}_{i}, B^{A}_{j})} - 1, 
        \end{equation}
        where $B^{A}$ represents the axis-aligned $B$, containing the top-left and bottom-right points $p\in\mathbb{R}^{1 \times 4}$ on the ground plane.
        Using the Bird's-Eye View (BEV) space as a case study, the alignment procedure is depicted in Fig. \ref{fig:f2}(d).
        The overlap and convex hull calculations in \cref{eq:agiou} are consistent with $gIoU_{2d}$~\cite{rezatofighi2019generalized}, with \textcolor{black}{$\mathcal{O}(n)$} computational complexity.
        \textcolor{black}{For clarity in subsequent discussions, we designate the parts of $gIoU$ and $A\text{-}gIoU$ excluding $IoU$ as the \textit{spatial occupancy difference}.}

        \textbf{\textcolor{black}{Delving into A-GIoU.}}
        % Naturally, two critical questions arise: 
        % (1) Can $A\text{-}gIoU$ effectively capture inter-object similarity?
        % Our answer is \textbf{affirmative}.
        % Moreover, incorporating $A\text{-}gIoU$ and scale-NMS~\cite{huang2021bevdet} in pre-processing even improves accuracy.
        % For a more comprehensive analysis, please refer to~\cref{Ablation}.
        As shown in Table~\ref{table:nu_metric_time} and~\ref{table:nu_ablation}, on nuScenes, integrating $A\text{-}gIoU$ reduces latency by 73\% while enhancing accuracy (+0.5\% MOTA). 
        This prompts the question: why does it excel in tracking tasks?
        \textcolor{black}{A critical insight is that $A\text{-}gIoU$ \textit{aligns closely with} $gIoU$ in describing \textit{the similarity between identical object pairs} and quantifying \textit{the discrepancy between different matching instances}.}
        
        \textcolor{black}{Take 3D space as an example, firstly, Fig.~\ref{fig:f2}(a) provides intuitive evidence of the similarity consistency between these two metrics for the same bbox pair across various geometric scenarios
        \footnote{\textcolor{black}{Both bboxes use nuScenes \textit{Car} average 3D size, with Z-position fixed at 1\text{m} and yaw angle sampled at $1.2^{\circ}$ step in $\in [0^{\circ}, 180^{\circ}]$.  
        X-Y positions: 
        Bbox1 at (0\text{m}, 0\text{m}); 
        Bbox2 is sampled at 1m step in $[-5\text{m}, 5\text{m}]$.}}.
        Generally, $A\text{-}gIoU$ slightly exceeds $gIoU$, with tiny differences in small object overlaps (distant or parallel bbox pairs) cases.
        To elucidate these phenomena, Fig.~\ref{fig:f2}(b, c) breaks down the distinctions between the two components of both metrics.
        Fig.~\ref{fig:f2}(c) reveals that \textit{spatial occupancy difference} is the key factor of this consistency.
        Fig.~\ref{fig:f2}(b) further highlights the source of the minor distinction and the enhancement provided by $A\text{-}gIoU$:
        It increases the $IoU$ for closely positioned bbox pairs, improving the tracker to recall observations with strong positional correlation despite inaccurate heading angles.}
        
        \textcolor{black}{Secondly, to validate the consistency between $A\text{-}gIoU$ and $gIoU$ in distinguishing true matches from false matches in data association, we introduce the Cost Gap Ratio (CGR) to quantify the performance disparity.
        CGR is defined as:
        \label{agiou}
        \begin{equation}
        \label{eq:CGR}
            CGR = (\min(C_{um}) - C_{m}) / (\max(C_{um}) - C_{m}), 
        \end{equation}
        where $C_{m}$ and $C_{um}$ are the cost of the true match and the false matches for a single tracklet
        \footnote{\textcolor{black}{Outliers ($C_{m} \geq \min(C_{um}$)) are excluded (representing $\le$ 3\% of data) to constrain CGR within the (0, 1] interval.}}.
        %, as calculated by the matching algorithm (Hungarian, Greedy, etc.).
        % Under the Hungarian algorithm setting, we retain matching instances where $C_{m}$ is less than $\min(C_{um})$, ensuring CGR falls within (0, 1), consistent with norms in local-optimal matching algorithms (Mutual Nearest Neighbor, Greedy, etc.).
        A higher CGR indicates a more pronounced difference between $C_{m}$ and $C_{um}$, suggesting a higher probability of a true positive match.
        % a denser contribution of $C_{um}$
        Fig.~\ref{fig:f2}(e) illustrates CGR values across all trajectories within randomly sampled scenes from nuScenes.
        The consistency between $A\text{-}gIoU$ and $gIoU$ in characterizing distinct matching instances is confirmed by similar Quartile1 (Q1), median, and Quartile3 (Q3) values.
        In addition, for $A\text{-}gIoU$, at least 75\% of CGR (Q1) exceeds 0.4 in the sampled scenes, ensuring the matching algorithm constructs accurate and stable bipartite graphs.}

        % From a single tracklet or detection perspective, due to the symmetry~\cite{rezatofighi2019generalized} of $gIoU$, the orientation error between $B^{A}$ and $B$ is $\le45^{\circ}$.
        % % (The detailed analysis is provided in \textcolor{red}{Appendix B}).
        % The limited angular loss indicates that $B^{A}$ can still characterize the spatial information of the object.
        % % For reference, on nuScenes, despite the mAOE (mean Average Orientation Error) of our baseline detector CenterPoint~\cite{yin2021center}, being $21^{\circ}$, the overall performance still achieves a high standard with a 60.3 mAP (mean Average Precision).
        % From the perspective of the correlation between detections and tracklets, Fig. \ref{fig:f2} demonstrates that the distances of matched pairs remain unchanged (or changed slightly), thereby illustrating how $A\text{-}gIoU$ works.
        % Intuitively, the alignment introduces a uniform scale offset on each trajectory and its corresponding observation.
        % Leveraging the size invariance~\cite{rezatofighi2019generalized} of $gIoU$, this consistent scale shift preserves the final similarity.
        % Consequently, the matched similarity remains notably higher than that of false matches.
        % Such consistency guarantees the Hungarian algorithm produces similar assignment structures compared to $gIoU$.

        \subsection{Densification}
        In this section, we improve the computational efficiency of the 3D tracking system through three key enhancements: 
        \textcolor{black}{soft tracklet elimination using tracking scores, 
        distant similarity calculation prevention via the voxel mask,
        and time-invariant states splitting with a lightweight filter.}

        \textbf{Confidence-Count Mixed Lifecycle.}
        \label{cb-life}
        Different from the rigid
        The classic count-based lifecycle seriously raises tracker maintenance costs.
        In contrast, \cite{benbarka2021score} flexibly terminates tracklets by manipulating scores, significantly reducing FP.
        However, as evidenced in \cite{benbarka2021score} and Table \ref{table:nu_life}, the deletion based on the latest score and the application of minus prediction detrimentally affect mismatch tracklets due to occlusion or FN detections.
        Towards this, a power function is first implemented to predict the scores smoothly.
        The refinement is described in two steps as follows:
        \begin{equation}
        \label{eq:refine}
        \begin{cases}
            \text{Predict: } s_{t, t-1} = \sigma \cdot s_{t-1}, \\
            \text{Update: } s_{t} = 1 - (1 - s_{t, t-1}) \cdot (1 - c_{t}), \\
        \end{cases}
        \end{equation}
        where $s_{t-1}$, $s_{t, t-1}$ and $s_{t}$ are the $t-1$ frame posterior score, $t$ frame priori score and $t$ frame posterior score of current tracklet, respectively.
        $c_{t}$ is the corresponding detection score.
        $\sigma$ is the hand-crafted decay rate, empirically $\le 0.7$.
        The update function is similar with~\cite{benbarka2021score}, satisfies that $s_{t}$ is more confidence than $s_{t, t-1}$ or $c_{t}$.
        After refinement, tracklets are terminated if their online average scores (computed from the initial frame to the current frame) fall below the deletion threshold $\theta_{dl}$ or if the number of mismatch frames exceeds the max-age.
        There are two noteworthy benefits:
        (1) Tracklets demonstrate increased resilience to temporary occlusions since their average confidence score decays gradually, as depicted in Table \ref{table:nu_life}.
        (2) The power function swiftly pulls the tracklet predict score to the $\left [ 0, \sigma   \right ]$ interval, leaving the over-confidence area ($\left [ 0.7, 1 \right ]$ interval in Fig. 2 of~\cite{benbarka2021score}). 
        This reduces the hysteresis in score estimates, improving the tracking performance.

        \textbf{Voxel Mask.} 
        In 3D MOT, an overlooked consideration is: 
        \textit{redundancy and matching are improbable between objects separated by significant distances in 3D Euclidean space.}
        This starkly contrasts inherently no-depth 2D MOT~\cite{liu2023fasttrack, zhou2020tracking}.
        % Most 3D MOT methods~\cite{pang2022simpletrack, li2023camo, li2023poly, weng20203d, wang2022deepfusionmot} devote substantial computational resources to calculating these ineffective costs. 
        To address this issue, we propose the voxel mask inspired by the coarse-to-fine idea~\cite{bai2022faster}.
        Specifically, before computing the global cost during data association and NMS iteration, we employ low-overhead Euclidean distance (2-norm) $||\cdot||_{2}$ and a hand-crafted size parameter $\theta_{vm}$ to generate voxel mask.
        The process can be expressed as:
        
        \begin{equation}
        \label{eq:voxel}
        \begin{cases}
            \text{Coarse: } M_{t} = ( ||\mathcal{B}^{xyz}_{1}-\mathcal{B}^{xyz}_{2}||_{2} \le \theta_{vm} ), \\
            \text{Fine: } C_{t} = 1 - \text{Aff}(M_{t}, \mathcal{B}_{1}, \mathcal{B}_{2}), \\
        \end{cases}
        \end{equation}
        where $\mathcal{B}$ is the 3D bounding boxes representing detections or tracklets based on module representation.
        $\mathcal{B}^{xyz}$ are the 3D positions.
        $M_{t}$ is the voxel mask between $\mathcal{B}_{1}$ and $\mathcal{B}_{2}$.
        \textcolor{black}{$\text{Aff}(\cdot)$ is the affinity function, constructing inter-object similarities between objects in pre-processing and association to detect redundancy and recall observations, respectively.}
        %\cref{eq:agiou} for tracklet-detection similarity in association, and $IoU_{bev}$ for inter-detection redundancy in pre-processing.}
        Notably, $\text{Aff}(\cdot)$ directly pads invalid values into false index within $M_{t}$ to expedite, ultimately yielding $C_{t}$.
        \label{voxel_mask}
        
        \textbf{Lightweight Filter.}
        \label{lwfilter}
        Current 3D MOT methods embed the time-invariant states (Z-position $z$, width $w$, length $l$, height $h$) of trajectory into the Kalman Filter.
        However, the time invariance makes the Jacobian matrices of process and observation identity matrices, resulting in numerous meaningless 0-related calculations.
        Besides, the computational time escalates swiftly with an expanding number of trajectories.
        To this end, we decouple these states $X^{ti}$ from the KF and estimate them using a compact filter. 
        The estimate process is expressed as:
        \begin{equation}
        \label{eq:median}
        \begin{cases}
            \text{Predict: } X^{ti}_{t, t-1} = X^{ti}_{t-1}, \\
            \text{Update: } X^{ti}_{t} = \text{LW-Filter}(X^{ti}_{t, t-1}, D^{i}_{t}), \\
        \end{cases}
        \end{equation}
        where $X^{ti}_{t-1}$, $X^{ti}_{t, t-1}$ and $X^{ti}_{t}$ are the $t-1$ frame posterior $X^{ti}$, $t$ frame priori $X^{ti}$ and $t$ frame posterior $X^{ti}$ of current tracklet, respectively.
        $D^{i}_{t}$ is the corresponding detection.
        LW-Filter($\cdot$) is the custom lightweight filter, implemented by a median or mean filter with length $l_{lw}$ in Fast-Poly.
        \textcolor{black}{It stores nearest fixed-size time-invariant trajectory observations for prompt state estimation.}
        There are two key insights: 
        (1) LW-Filter($\cdot$) bypasses invalid computations, enabling fast filtering while preserving temporal data.
        (2) In contrast to KF, which entails manual process and observation noise, LW-Filter($\cdot$) relies solely on a single hyperparameter $l_{lw}$, reducing overfitting and ensuring robustness, as proven in Fig. \ref{fig:f4}.
        
        % \begin{algorithm}[t]
        % \label{algo}
        % \SetAlgoLined
        % \DontPrintSemicolon
        % \SetNoFillComment
        % \footnotesize
        % \SetKwIF{Wait}{}{}{wait}{then}{}{}{end}
        % \KwIn{Current 3D raw detections $D_t'$; Previous alive trajectories $T_{t-1}$; Frame number $N_f$ }
        % \KwOut{Alive trajectories at current frame $T_t$}
        % Initialization: Two parallel processes $P_1, P_2$; Shared flag $Flag = 1$; Shared variables $V = \emptyset$\;
        % \# Process $P_1$: Pre-processing raw detections of each frame $t$\;
        % \For{$t$ $\leftarrow$ $0$ \KwTo $N_f$}{
        %   \Wait {$Flag==1$}{
        %   $D_t$ $\leftarrow$ \texttt{Pre-processing}($D_t'$)\;
        %     $V$ $\leftarrow$ $D_t$;
        %     $Flag$ $\leftarrow$ $0$
        %   }
        % } 
        % \# Process $P_2$: Predict trajectories and implement online tracking\;
        % \For{$t$ $\leftarrow$ $0$ \KwTo $N_f$}{
        % $T_{t,t-1}$ $\leftarrow$ \texttt{Predict}($T_{t-1}$)\;
        %   \Wait {$Flag == 0$}{
        %   $T_t$ $\leftarrow$ \texttt{Tracking}($T_{t,t-1}, V(D_t)$)\;
        %   $V$ $\leftarrow$ $\emptyset$;
        %   $Flag$ $\leftarrow$ $1$
        %   }
        % }
        % Return: $T_t, t = 0, 1, ..., N_f$
        % \caption{Pseudo-code of Parallelization}
        % \label{alg:algorithm1}
        % \end{algorithm}
        
        \subsection{Parallelization}
        \label{parallel}
        Although the TBD framework is inherently serial, certain modules can be parallelized.
        Consequently, in Fast-Poly, after tracking in the previous frame, motion estimation prediction and preprocessing are synchronized using multi-processing technology.
        % Consequently, in Fast-Poly, following the tracking in the previous frame, the prediction in the motion estimation and the preprocessing are synchronized based on the multi-processing technology.
        % This process is illustrated in the pseudo-code provided in \cref{alg:algorithm1}. 
        % This process is provided as a pseudo-code in \cref{alg:algorithm1}. 
        Specifically, before the online tracking, two parallel processes $P_1, P_2$ are initiated.
        This initialization also includes a shared flag $Flag$ and variable $V$ to synchronize status and transmit data between processes. 
        At each frame $t$, $P_1$ filters raw detections while $P_2$ concurrently predicts the previous alive trajectory and implements online tracking.
        % Upon acquiring preprocessing results, $P_1$ enters a sleep state, allowing $P_2$ to commence tracking. 
        % Following the completion of tracking, $P_2$ awakens $P_1$, enabling both processes to concurrently handle the next frame.
        After completing their respective tasks, $P_1$ and the prediction part of $P_2$ enter a sleep state, remaining dormant until they are simultaneously awakened in the subsequent frame.
        Fast-Poly adheres to online tracking principles and fundamentally differs from~\cite{pang2022simpletrack}, which employs multi-processing for synchronous multi-scene tracking.

	\section{Experiments}
        \label{Experiments}
	\subsection{Dataset and Implementation Details}
	\textbf{nuScenes.}
        nuScenes~\cite{caesar2020nuscenes} provides intricate scenes and numerous short sequences (40 frames) with 2Hz for keyframe data.
        Evaluation encompasses 7 categories: Car (\textit{Car}), Bicycle (\textit{Bic}), Motorcycle (\textit{Moto}), Pedestrian (\textit{Ped}), Bus (\textit{Bus}), Trailer (\textit{Tra}), and Truck (\textit{Tru}).
        The primary evaluation metric is AMOTA~\cite{weng20203d}.
        Fast-Poly is conducted utilizing only keyframes.
        
        \textbf{Waymo.} 
        Waymo~\cite{sun2020scalability} provides extended sequences (approximately 20 seconds each) for tracking Vehicle (\textit{Veh}), Cyclist (\textit{Cyc}), and Pedestrian (\textit{Ped}), captured at a frequency of 10Hz. 
        The main evaluation metric, MOTA~\cite{bernardin2006multiple}, is categorized into two levels based on the LiDAR points in labels: LEVEL\_1 and LEVEL\_2. 
        MOTA on LEVEL\_2 is reported due to the cumulative nature of difficulty levels.
        
        % \textbf{KITTI.}
        % KITTI~\cite{geiger2012we} requires tracking of \textit{Car} or \textit{Ped} on long sequences with a ground truth frequency of 10Hz.
        % HOTA~\cite{luiten2021hota} serves as the primary metric.
        % Notably, as a 2D tracking benchmark, the projection boxes of the 3D tracking results are eventually evaluated.

        \begin{figure}[t]
        \centering
        %\framebox{\parbox{3in}{We suggest that you use a text box to insert a graphic (which is ideally a 300 dpi TIFF or EPS file, with all fonts embedded) because, in a document, this method is somewhat more stable than directly inserting a picture.}}
        \includegraphics[width=\linewidth]{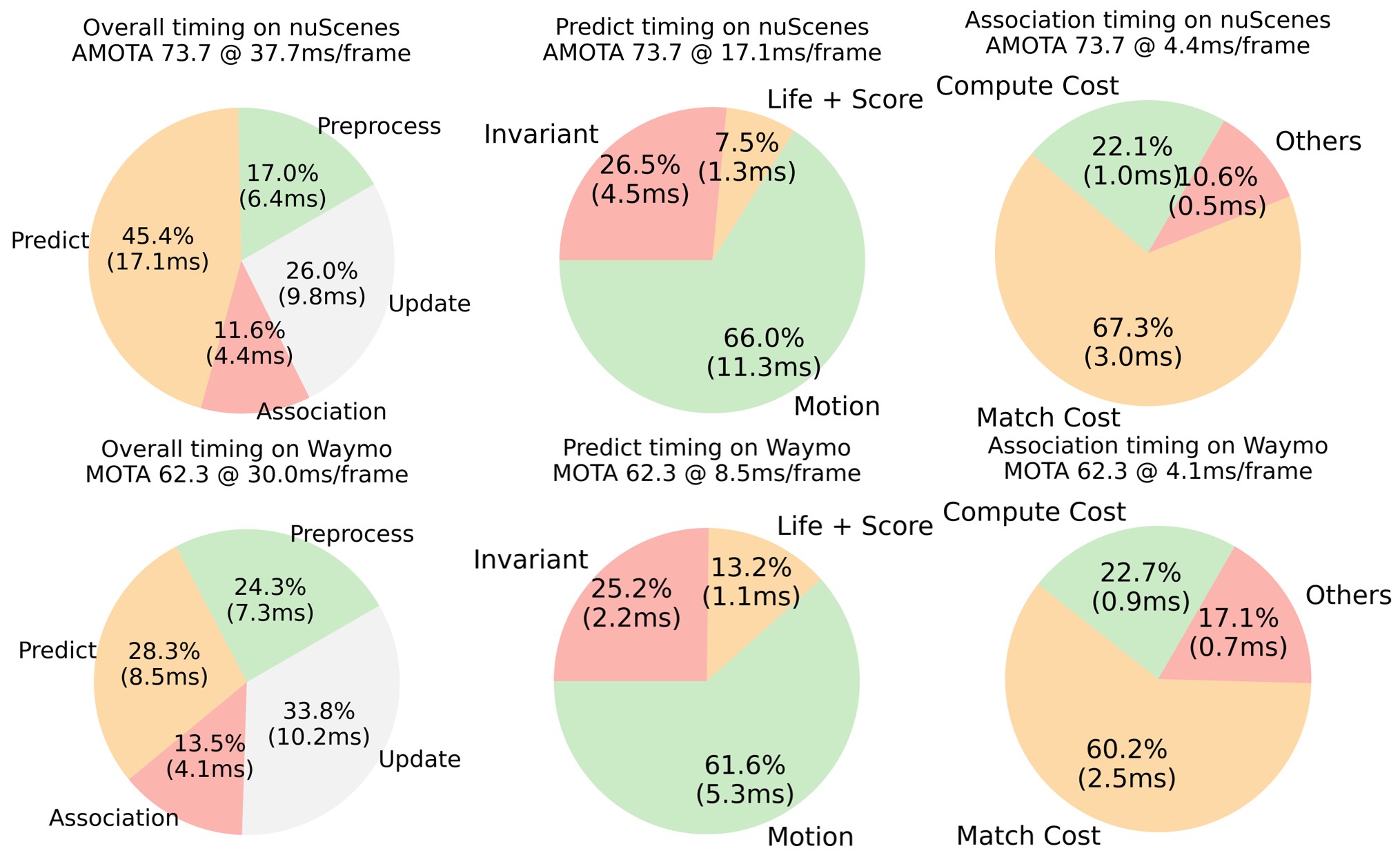}
        \caption{The average timing statistics of each module in Fast-Poly on the nuScenes and Waymo val set without parallelization.
        \textbf{Invariant} means the lightweight filter for time-invariant states.
        \textbf{Motion} means the Kalman Filter for motion states.
        \textbf{Life} means the lifecycle module.
        \textbf{Score} means the score refinement.
        % Notably, the parallelization technology will directly cut out the more time-consuming modules of pre-processing and prediction.
        }
        \label{fig:f3}
        \end{figure}
        
        \textbf{Implementation Details.}
        % input part
        % For a fair comparison, we employ the baseline detectors CenterPoint~\cite{yin2021center} and LargeKernel3D~\cite{chen2022scaling} on nuScenes validation and test set, respectively.
        % device and language
        \label{implementation}
        Under Python implementation with NumPy~\cite{van2011numpy}, Fast-Poly is deployed and tested.
        % hyperparameters find
        We perform the linear search to refine the optimal AMOTA (nuScenes) and MOTA (Waymo) hyperparameters on the validation set. 
        The best hyperparameters are subsequently utilized on the test set.
        % pre-processing: score filter
        SF thresholds are category-specific and detector-specific, which are (\textit{Moto, Car, Bic}: 0.16; \textit{Bus, Tra}: 0.13; \textit{Tru}: 0; \textit{Ped}: 0.19) on nuScenes, (\textit{Veh}: 0.8; \textit{Ped}: 0.3; \textit{Cyc}: 0.84) on Waymo.
        % pre-processing: NMS
        The NMS thresholds are 0.08 on all categories and datasets.
        % We also employ Scale-NMS~\cite{huang2021bevdet} on (\textit{Bic}, \textit{Ped}) on nuScenes and \textit{Bic} on Waymo.
        With default $IoU_{bev}$ in NMS, we additionally utilize our proposed $A\text{-}gIoU_{bev}$ to describe similarity for (\textit{Bic}, \textit{Ped}, \textit{Bus}, \textit{Tru}) on nuScenes, (\textit{Veh, \textit{Ped}, \textit{Cyc}}) on Waymo.
        % estimation
        The motion models and filters are consistent with~\cite{li2023poly}.
        The lightweight filter is implemented by the median filter with $l_{lw}=3$ on all datasets.
        % association
        The association metrics are all implemented by $A\text{-}gIoU$ on all datasets.
        The first association thresholds $\theta_{fm}$ are category-specific, which are (\textit{Bic, Moto}: 1.6; \textit{Bus, Car, Tra, Tru}: 1.2; \textit{Ped}: 1.8) on nuScenes, (\textit{Veh}: 1.1; \textit{Ped}: 1.3; \textit{Cyc}: 1.2) on Waymo.
        Voxel mask size $\theta_{vm}$ is 3\textit{m} on nuScenes and 5\textit{m} on Waymo. 
        % lifecycle: count-based and output
        The count-based and output file strategies are consistent with~\cite{li2023poly}.
        % lifecycle: confidence-based
        In the confidence-based part, decay rates $\sigma$ are category-specific, which are (\textit{Moto, Ped}: 0.6; \textit{Car, Tra}: 0.5; \textit{Tru}: 0.2; \textit{Bus}: 0.3; \textit{Bic}: 0.1) on nuScenes, (\textit{Veh}: 0.6; \textit{Ped}: 0.7; \textit{Cyc}: 0.1) on Waymo.
        The delete threshold $\theta_{dl}$ are (\textit{Bus, Ped}: 0.1 and 0.04 for other categories) on nuScenes, (\textit{Cyc}: 0.2; \textit{Veh, Ped}: 0.1) on Waymo.
        
        % metric
        \textbf{Metrics.}
        We present \textcolor{black}{three} primary indicators: MOTA~\cite{bernardin2006multiple}, AMOTA~\cite{weng20203d}, and FPS (Frame Per Second), supplemented by three secondary indicators: IDS (ID Switch), FP, and FN.
        The latency of each experiment is the average time cost under multiple evaluations.
        \textcolor{black}{We only measure the tracking time for all methods, excluding the detection time.}

        \begin{table*}[t]
        \centering
        \caption{A comparison between our proposed method with other advanced methods on the nuScenes test set.
        This leaderboard is available at \href{https://www.nuscenes.org/tracking?externalData=all&mapData=all&modalities=Any}{nuScenes official benchmark}.
        $\ddagger$ means the GPU device.
        \textcolor{black}{The reported runtimes of all methods exclude the detection time.}
        Fast-Poly and~\cite{li2023poly} rely entirely on the detector input, as they do not utilize any visual or deep features \textcolor{black}{during tracking}.}
        \label{table:nu_test}
        \renewcommand{\arraystretch}{0.7}
        \setlength{\tabcolsep}{2.3mm}{
        \begin{tabular}{cccc|ccc|ccc}
        \toprule
        \multicolumn{1}{c}{\textbf{Method}} & \textbf{Device} & \textbf{Detector} & \textbf{Input} & \textbf{AMOTA}$\uparrow$ & \textbf{MOTA}$\uparrow$ & \textbf{FPS}$\uparrow$ & \textbf{IDS}$\downarrow$ & \textbf{FN}$\downarrow$ & \textbf{FP}$\downarrow$ \\ \midrule
                                     EagerMOT~\cite{kim2021eagermot}               & \textbf{\text{--}}       & CenterPoint~\cite{yin2021center}\&Cascade R-CNN~\cite{cai2018cascade}         & 2D+3D       & 67.7      & 56.8     & 4    & 1156    & 24925   & 17705   \\
                                   CBMOT~\cite{benbarka2021score}     & I7-9700          & CenterPoint~\cite{yin2021center}\&CenterTrack~\cite{zhou2020tracking}          & 2D+3D         & 67.6      & 53.9      & \textbf{80.5}    & 709    & 22828   & 21604   \\
                                   % ShaSTA~\cite{sadjadpour2023shasta} & A100$\ddagger$       & CenterPoint~\cite{yin2021center}         & 3D      & 69.6      & 57.8     & 10     & 473    & 21293   & 16746   \\
                                   Minkowski~\cite{gwak2022minkowski} & TITAN$\ddagger$  & Minkowski~\cite{gwak2022minkowski}         & 3D      & 69.8      & 57.8     & 3.5    & 325    & 21200   & 19340   \\
                                   ByteTrackv2~\cite{zhang2023bytetrackv2} & \textbf{\text{--}}       & TransFusion-L~\cite{bai2022transfusion}         & 3D      & 70.1      & 58     & \textbf{\text{--}}    & 488    & 21836   & 18682   \\
                                   3DMOTFormer~\cite{ding20233dmotformer}& 2080Ti$\ddagger$       & BEVFuison~\cite{liu2023bevfusion}       & 2D+3D      & 72.5      & 60.9     & 54.7    & 593    & 20996   & 17530   \\  
                                   %CAMO-MOT~\cite{li2023camo}& 3090Ti$\ddagger$   & BEVFuison~\cite{liu2023bevfusion}\&FocalsConv~\cite{chen2022focal}   & 2D+3D      & 75.3      & \textbf{63.5}     & \textbf{\text{--}}    & 324    & 18192   & 17269   \\
                                   Poly-MOT~\cite{li2023poly}               & 9940X       & LargeKernel3D~\cite{chen2022scaling}         &2D+3D       & 75.4      & 62.1     & 3    & \textbf{292}    & \textbf{17956}   & 19673   \\ \midrule
                                  \textbf{Fast-Poly (Ours)}               & 7945HX       & LargeKernel3D~\cite{chen2022scaling}         &2D+3D       & \textbf{75.8}      & 62.8     & 34.2    & 326    & 18415   &  \textbf{17098}  \\
        \bottomrule
        \end{tabular}}
        \vspace{-1.5em}
        \end{table*}

        \begin{table}[t]
        \centering
        \caption{A comparison of the average time consumption of each step of GIoU and our proposed metric A-GIoU in data association on nuScenes val set.
        \textbf{Overlap} means solving intersections.
        \textbf{Convex} means solving convex hulls.
        }
        \label{table:nu_metric_time}
        \renewcommand{\arraystretch}{0.7}
        \begin{tabular}{ccccc}
        \toprule
        \multicolumn{1}{c}{\textbf{Metric}} & \textbf{Overall} & \textbf{Overlap} & \textbf{Convex} & \textbf{Others}\\ \midrule
        $gIoU$                            & 111.3ms      & 24.7ms     & 83.0ms  & 3.6ms    \\
        ${A\text{-}gIoU}$      & 0.0098ms     & 0.0033ms  & 0.0016ms  & 0.0049ms \\
        \bottomrule
        \end{tabular}
        \vspace{-1em}
        \end{table}

        \begin{table}[t]
        \centering
        \caption{A comparison of the \textcolor{black}{run}-time performance on different devices on nuScenes val set with CenterPoint.
        }
        \label{table:nu_device}
        \renewcommand{\arraystretch}{0.7}
        \begin{tabular}{ccccc}
        \toprule
        \multicolumn{1}{c}{\textbf{Device}} & \textbf{Level} & \textbf{FPS}$\uparrow$ & \textbf{AMOTA}$\uparrow$ & \textbf{MOTA}$\uparrow$ \\ \midrule
         Ryzen9 7945HX            & Personal      & 28.9     & \multirow{3}{*}{73.7}   & \multirow{3}{*}{63.2}    \\ %all 208.73s
         Intel 9940X       & Server      & 13.4      &   &    \\ 
         Cortex A78AE      & Embedded      & 8.7     &   &    \\ %all 693s
        \bottomrule
        \end{tabular}
        \end{table}

        % \begin{table}[t]
        % \centering
        % \caption{A comparison between our proposed method with other advanced methods on the Waymo val set.
        % }
        % \label{table:waymo_val}
        % \renewcommand{\arraystretch}{1.3}
        % \setlength{\tabcolsep}{2.7mm}{
        % \begin{tabular}{ccc|c}
        % \toprule
        % \multicolumn{1}{c}{\textbf{Method}} & \textbf{Detector} & \textbf{Input}  & \textbf{MOTA}$\uparrow$ \\ \midrule
        %                          %     &   &   &       \\
        %                      CenterPoint~\cite{yin2021center}        & CenterPoint~\cite{yin2021center} & 3D &  55.8 \\
        %                      SimpleTrack~\cite{pang2022simpletrack}        & CenterPoint~\cite{yin2021center} & 3D &  56.9 \\
        %                      ImmotralTrack~\cite{wang2021immortal}& CenterPoint~\cite{yin2021center} & 3D & 57.9 \\
        %                         CasTrack~\cite{wu20213d}        & CasA~\cite{wu2022casa} & 3D & 61.3 \\ \hline
        %                        \textbf{Fast-Poly (Ours)}        & CasA~\cite{wu2022casa} & 3D &    \\
        % \bottomrule
        % \end{tabular}}
        % \end{table}
        
        \begin{table}[t]
        \centering
        \caption{A comparison between our proposed method with other advanced methods on the Waymo val and test set.
        The leaderboard is available at \href{https://waymo.com/open/challenges/2020/3d-tracking/}{Waymo official benchmark}.
        The inference speed is 35.5 FPS.
        }
        \label{table:waymo_test_val}
        \renewcommand{\arraystretch}{0.8}
        \setlength{\tabcolsep}{2mm}{
        \begin{tabular}{ccc|c|c}
        \toprule
        \multicolumn{1}{c}{\textbf{Split}} & \textbf{Method} & \textbf{Detector} & \textbf{Input}  & \textbf{MOTA}$\uparrow$ \\ \midrule
                       \multirow{4}{*}{Val}     &CenterPoint~\cite{yin2021center}        & CenterPoint~\cite{yin2021center} & \multirow{4}{*}{3D} &  55.8 \\
                             &SimpleTrack~\cite{pang2022simpletrack}        & CenterPoint~\cite{yin2021center} &  &  56.9 \\
                            % & ImmotralTrack~\cite{wang2021immortal}& CenterPoint~\cite{yin2021center} &  & 57.9 \\
                               & CasTrack~\cite{wu20213d}        & CasA~\cite{wu2022casa} &  & 61.3 \\ 
                            &   \textbf{Fast-Poly (Ours)}        & CasA~\cite{wu2022casa} &  & \textbf{62.3}   \\  \hline
                           \multirow{5}{*}{Test}   & PVRCNN-KF~\cite{shi2020pv}      &  PVRCNN~\cite{shi2020pv} & \multirow{5}{*}{3D}  &  55.5     \\
                             % AlphaTrack~\cite{zeng2021cross}         &AlphaTrack~\cite{zeng2021cross}          &  2D+3D  & 57.4  \\
                             & CenterPoint~\cite{yin2021center}        & CenterPoint~\cite{yin2021center} &  &  58.7 \\
                             & SimpleTrack~\cite{pang2022simpletrack}        & CenterPoint~\cite{yin2021center} &  &  60.2 \\
                             % & ImmotralTrack~\cite{wang2021immortal}& CenterPoint~\cite{yin2021center} &  & 60.9  \\
                               &  CasTrack~\cite{wu20213d}        & CasA~\cite{wu2022casa} &  & 62.6 \\ 
                              & \textbf{Fast-Poly (Ours)}        & CasA~\cite{wu2022casa} &  & \textbf{63.6}   \\
        \bottomrule
        \end{tabular}}
        \end{table}

        \begin{table*}[t]
        \centering
        \caption{The ablation studies of each module on the nuScenes val.
        \textbf{AG} means A-GIoU.
        \textbf{VM} means Voxel Mask.
        \textbf{LW} means LightWeight Filter.
        \textbf{CL} means Confidence-Count Mixed Lifecycle.
        \textbf{MP} means Multi-Processing.
        }
        \label{table:nu_ablation}
        \renewcommand{\arraystretch}{0.7}
        \setlength{\tabcolsep}{1.8mm}{
        \begin{tabular}{cccccc|ccc|ccc}
        \toprule
        \multirow{2.5}{*}{\textbf{Index}} & \textbf{Alignment} & \multicolumn{3}{c}{\textbf{Densification}} & \textbf{Parallelization}  & \multicolumn{3}{c}{\textbf{Primarily Metric}} & \multicolumn{3}{c}{\textbf{Secondary Metric}} \\ 
        \cmidrule(lr){2-2} \cmidrule(lr){3-5} \cmidrule(lr){6-6} \cmidrule(lr){7-9} \cmidrule(lr){10-12}
                               & \textbf{AG}    & \textbf{VM}        & \textbf{LW}        & \textbf{CL}        & \textbf{MP} &\textbf{AMOTA}$\uparrow$ & \textbf{MOTA}$\uparrow$ & \textbf{FPS}$\uparrow$ &\textbf{IDS}$\downarrow$ & \textbf{FP}$\downarrow$ & \textbf{FN}$\downarrow$  \\ \midrule
        \textbf{Baseline}~\cite{li2023poly}               & \textbf{\text{--}}          & \textbf{\text{--}}          & \textbf{\text{--}}          & \textbf{\text{--}}          & \textbf{\text{--}}                 & 73.1                      & 61.9                      & 5.6 (188.7ms)                     & \textbf{232}                     & 13051                    & 17593                \\
        \textbf{Exp1}                   & \checkmark           & \textbf{\text{--}}          & \textbf{\text{--}}          & \textbf{\text{--}}          & \textbf{\text{--}}                 & 73.1                       & 62.4                      & 19.9 (50.3ms)                     & 342                     & 13955                    & 16794              \\
        \textbf{Exp2}                   & \checkmark          & \checkmark          & \textbf{\text{--}}          & \textbf{\text{--}}          & \textbf{\text{--}}                 & 73.3                      & 63.0                      & 21.3 (46.9ms)                     & 385                     & \textbf{12732}                    & 17909                   \\
        \textbf{Exp3}                   & \checkmark          & \checkmark          & \checkmark          & \textbf{\text{--}}          & \textbf{\text{--}}                 & 73.3                     & 62.9                      & 21.7 (46.1ms)                    & 427                     & 14124                    & 16523                 \\
        \textbf{Exp4}                   & \checkmark          & \checkmark          & \checkmark          & \checkmark          & \textbf{\text{--}}                 & 73.7                         & 63.2                      & 26.5 (37.7ms)                    & 414                     & 14713                    & 15900                  \\ %227s
        \textbf{Exp5}                   & \checkmark           & \checkmark          & \checkmark          & \checkmark          & \checkmark                 &  \textbf{73.7}                      &  \textbf{63.2}                     &  \textbf{28.9 (34.6ms)}                   &  414                    & 14713                    & \textbf{15900}                   \\
        \bottomrule
        \end{tabular}}
        \vspace{-1.5em}
        \end{table*}

        \begin{table}[t]
        \centering
        \caption{A comparison on distinct lifecycle modules on nuScenes val set.
        \textbf{Average} means using the online average score to delete.
        \textbf{Latest} means using the latest score to delete.
        \textbf{Max-age} means using the continuous mismatch time to delete.
        Other settings are under the best performance.
        }
        \label{table:nu_life}
        \renewcommand{\arraystretch}{0.7}
        \begin{tabular}{ccccc}
        \toprule
        \multicolumn{1}{c}{\textbf{Strategy}} & \textbf{AMOTA}$\uparrow$ & \textbf{MOTA}$\uparrow$ & \textbf{FPS}$\uparrow$ & \textbf{FN}$\downarrow$\\ \midrule
         Count                \& Max-age            & 73.3      & 62.9     & 23.0  & 16523    \\
         %predict:normal update:normal+delete = -100
         Confidence~\cite{benbarka2021score} \& Latest       & 70.6      & 63.2     & \textbf{45.8}  & 19192   \\%predict:minus update:multi+latest
         Confidence~\cite{benbarka2021score} \& Average      & 73.3      & 63.1     & 28.3 & 16826   \\%predict:minus update:multi+average
         Confidence (Ours)     \& Average      & \textbf{73.7}      & \textbf{63.2}     & 28.9  & \textbf{15900}   \\%predict:normal update:multi+average 27.7or28.9
        \bottomrule
        \end{tabular}
        \end{table}

        \subsection{Run-time Discussion}

        \textbf{The comparison with other methods.}
        As shown in Tables \ref{table:nu_test} and \ref{table:waymo_test_val}, our inference speed is extremely competitive on both datasets.
        % compared to baseline Poly-MOT
        We outline the detailed time consumption of each module in Figure \ref{fig:f3} and Table~\ref{table:nu_metric_time}.
        On nuScenes, Fast-Poly can run at 34.2 FPS without any GPU.
        Besides, the association timing is significantly reduced to 4.4ms, which accounts for only 4\% of the original $gIoU$ calculation time.
        Due to the higher-quality detections (due to the differing requirements between MOTA and AMOTA), the overall timing (35.5 FPS) as well as that of each module is further reduced for the Waymo dataset.
        % limitation
        The estimation module dominates the latency, constituting 60\% of the total, largely due to the computational demands of solving the Jacobian matrix in the EKF.
        
        \textbf{The robustness of device migration.}
        As demonstrated in Table \ref{table:nu_device}, we evaluated Fast-Poly across three CPU tiers. 
        Since our method is learning-free, tracking accuracy remains unaffected by device migration.
        On the personal-level CPU, Fast-Poly exhibits the best real-time performance (29 FPS).
        \textcolor{black}{Furthermore, without engineering acceleration, Fast-Poly can achieve nearly real-time performance (9 FPS) on limited CPU resources (Jetson Orin AGX), highlighting its practicality.}
        % Besides, considering the challenging outdoor sequences of nuScenes, the inference speed (9 FPS) on Jetson Orin AGX enables its deployment in practical robotic systems.

        % \begin{figure}[t]
        % \centering
        % \includegraphics[width=0.7\linewidth]{./7.png}
        % \caption{
        % The time consumption curve of average frame timing with the objects increasing between Fast-Poly and baseline method Poly-MOT on nuScenes val set.
        % }
        % \label{fig:f7}
        % \end{figure}
        
        \textbf{The parallel efficiency of Fast-Poly.}
        As depicted in Fig. \ref{fig:f0}, when the number of objects increases significantly, the frame timing of Fast-Poly exhibits minimal increments ($\le$30ms), highlighting the outstanding computational efficiency and feasibility of our method in autonomous driving scenarios.
        
        \begin{table}[t]
        \centering
        \caption{A comparison between our proposed method with other advanced methods on the nuScenes val set.
        $\ddagger$ means the GPU device.
        All metrics in the competition papers are reported.
        % Methods in lines 1-7 and lines 8-9 use CenterPoint~\cite{yin2021center} and LargekKernel3D-L~\cite{chen2022scaling}, respectively.
        \textcolor{black}{Methods in lines 1-7, lines 8-9, and line 10-13 use CenterPoint~\cite{yin2021center}, LargeKernel3D~\cite{chen2022scaling} and DETR3D~\cite{DETR3D}.}
        }
        \label{table:nu_val}
        \renewcommand{\arraystretch}{0.7}
        \setlength{\tabcolsep}{1.6mm}{
        \begin{tabular}{cc|cccc}
        \toprule
        \multicolumn{1}{c}{\textbf{Method}} & \textbf{Device} & \textbf{AMOTA}$\uparrow$ & \textbf{MOTA}$\uparrow$ & \textbf{FPS}$\uparrow$ & \textbf{IDS}$\downarrow$  \\ \midrule
                    OGR3MOT~\cite{zaech2022learnable}               & TITANXp$\ddagger$        & 69.3      & 60.2     & 12.3    & 262       \\
                    SimpleTrack~\cite{pang2022simpletrack}               & 9940X      & 69.6      & 60.2     & 0.5    & 405       \\
                    3DMOTFormer~\cite{ding20233dmotformer}               &  2080Ti$\ddagger$      & 71.2      & 60.7     & \textbf{54.7}    & 341      \\
                    ByteTrackv2~\cite{zhang2023bytetrackv2}               & \textbf{\text{--}}      & 72.4      & 62.4     & \textbf{\text{--}}    & \textbf{183}       \\
                    ShaSTA~\cite{sadjadpour2023shasta} & A100$\ddagger$      & 72.8      & \textbf{\text{--}}     & 10.0     & \textbf{\text{--}}      \\
                    Poly-MOT~\cite{li2023poly}               & 7945HX      & 73.1      & 61.9     & 5.6    & 232       \\
                    \textbf{Fast-Poly (Ours)}               &  7945HX      & \textbf{73.7}      & \textbf{63.2}     & 28.9    & 414       \\ \midrule
                    Poly-MOT~\cite{li2023poly}               & 7945HX  & 75.2      & 54.1    & 8.6    & \textbf{252}       \\
                    \textbf{Fast-Poly (Ours)}               &  7945HX  & \textbf{76.0}      & \textbf{65.8}     & \textbf{34.2}    & 307      \\
        
            \midrule
            % \textcolor{blue}{StreamPETR~\cite{streampetr}}             & \textcolor{blue}{57.88} & \textcolor{blue}{StreamPETR~\cite{streampetr}} & \textcolor{blue}{42.2} & \textcolor{blue}{1.185} & \textcolor{blue}{785} & \textcolor{blue}{12537} & \textcolor{blue}{41461} \\
                \textcolor{black}{CC-3DT~\cite{cc3dt}}     &  \textcolor{black}{ 3090$\ddagger$} & \textcolor{black}{35.9}  & \textcolor{black}{32.6} &  \textcolor{black}{9.6} & \textcolor{black}{2152}  \\

                \textcolor{black}{PF-Track~\cite{pftrack}}     &  \textcolor{black}{A100$\ddagger$} & \textcolor{black}{36.2}  & \textcolor{black}{\textbf{\text{--}}} &  \textcolor{black}{\textbf{\text{--}}} & \textcolor{black}{\textbf{300}}  \\

                % \textcolor{blue}{ByteTrackv2~\cite{zhang2023bytetrackv2}}     &  \textcolor{blue}{V100$\ddagger$} & \textcolor{blue}{35.9}  & \textcolor{blue}{36.2} &  \textcolor{blue}{\textbf{18.2}} & \textcolor{blue}{2152}  \\
                
                % \textcolor{blue}{ADA-Track~\cite{ding2024ada}}     &  \textcolor{blue}{2080Ti$\ddagger$} & \textcolor{blue}{39.2}  & \textcolor{blue}{36.3} &  \textcolor{blue}{\textbf{3.2}} & \textcolor{blue}{897}  \\
                
                \textcolor{black}{Poly-MOT~\cite{li2023poly}}               & \textcolor{black}{7945HX}  & \textcolor{black}{38.7}      & \textcolor{black}{32.3}    & \textcolor{black}{16.3}    & \textcolor{black}{926}       \\
                
                \textcolor{black}{\textbf{Fast-Poly (Ours)}}               &  \textcolor{black}{7945HX}  & \textcolor{black}{\textbf{40.7}}      & \textcolor{black}{\textbf{33.1}}     & \textcolor{black}{\textbf{39.3}}    & \textcolor{black}{1312}      \\
        \bottomrule
        \end{tabular}}
        \end{table}

        \subsection{Comparative Evaluation}
        
        \textbf{nuScenes.}
        % test
        Fast-Poly establishes a new state-of-the-art performance on the test set, with 75.8\% AMOTA and 34.2 FPS, surpassing all existing methods.
        With the same detector, Fast-Poly outperforms Poly-MOT~\cite{li2023poly} on all major metrics (accuracy: +0.4\% AMOTA and +0.7\% MOTA, latency: +31.2 FPS).
        Despite being marginally slower than CBMOT~\cite{benbarka2021score} and 3DMOTFormer~\cite{ding20233dmotformer} in speed, Fast-Poly significantly outperforms them in accuracy.
        On a limited computing platform (only personal CPU), Fast-Poly maintains high accuracy and high real-time consistency.
        Our method is open-sourced, serving as a strong baseline for 3D MOT.

        % val
        On the val set, we employ CenterPoint~\cite{yin2021center} as the detector for a fair comparison.
        As shown in Table \ref{table:nu_val}, Fast-Poly outperforms most learning-based/free methods by a significant margin in terms of both tracking accuracy (73.7\% AMOTA, 64.2\% MOTA) and latency (29 FPS).
        We surpass the baseline~\cite{li2023poly} with a notable improvement of +1.3\% in MOTA and +0.6\% in AMOTA accuracy, while 5$\times$ faster under identical settings.
        Fast-Poly exhibits stronger tracking performance with a powerful detector LargeKernel3D~\cite{chen2022scaling}.
        \textcolor{black}{Utilizing the multi-camera detector DETR3D~\cite{DETR3D} with constrained performance, Fast-Poly shows robust real-time performance and accuracy, exceeding the end-to-end tracker PF-Track (+4.5\% AMOTA) and TBD tracker CC-3DT~\cite{cc3dt} (+4.8\% AMOTA).
        Notably, Poly-MOT requires a strict score filter threshold to achieve optimal AMOTA, reducing the latency advantage of our method.}

        \textbf{Waymo.}
        %test
        As shown in Table ~\ref{table:waymo_test_val}, with the same detector, we exceed CasTrack~\cite{PC3T} with a +1\% improvement in MOTA on both sets, showcasing superior tracking performance.
        Without using the state-of-the-art detector, Fast-Poly ranks 4th on the Waymo tracking leaderboard among all LiDAR-only online methods with 63.6\% MOTA to date.
        Despite the scarcity of methods showcasing real-time performance on Waymo, we believe that our inference speed (35.5 FPS) is competitive.
        %val
        % On the val set, Fast-Poly outperforms other trackers in terms of higher MOTA when using the same detector(CasA). We exceeded the baseline, showcasing a notable enhancement of +XX in MOTA accuracy. 

        \subsection{Ablation Studies}
        \label{Ablation}

        To verify the effectiveness of each module, we conducted comprehensive ablation experiments on the nuScenes val set.
        Our baseline tracker Poly-MOT~\cite{li2023poly} is re-implemented with the official \href{https://github.com/lixiaoyu2000/Poly-MOT}{code}.
        With baseline detector CenterPoint~\cite{yin2021center}, the best performance for each ablation experiment is reported.

        \textbf{The effect of the Proposed Metric A-GIoU.}
        As demonstrated in Tables \ref{table:nu_metric_time} and \ref{table:nu_ablation}, $A\text{-}gIoU$ results in a significant reduction in latency (-138.4ms), particularly during the overlap (-24.7ms) and convex hull (-83ms) calculations.
        \textcolor{black}{$A\text{-}gIoU$ maintains AMOTA (73.1\%) while improving MOTA (+0.5\%).
        Fig.~\ref{fig:f2}(a) shows this refinement stems from focusing on tight positional correlations, recalling more observations (-799 FN).}
        % \textcolor{blue}{This will make the algorithm more inclined to weaken the erroneous angle estimates generated during detection and prediction and focus on the positional correlation between the detection and prediction frames. 
        % Of course, this affinity exaggeration makes correlating between the two sets(+951 FP and -799 FN) easier. Still, the final result proves that $A\text{-}gIoU$ can speed up the computation without compromising the tracking accuracy(73.1\% AMOTA and +0.5\% MOTA).}
        % Moreover, as illustrated in Table~\ref{table:nu_pre}, the combination of $A\text{-}gIoU$ and Scale-NMS~\cite{huang2021bevdet} leads to a reduction in FP in small categories (\textit{Bic}: -52 FP) and a boost in accuracy (+0.1\% AMOTA).
        % This is because $A\text{-}gIoU$ can more comprehensively consider the spatial distance difference between objects than $IoU$.
        
        \textbf{The effect of the Voxel Mask.}
        As illustrated in Table \ref{table:nu_ablation}, the voxel mask improves overall accuracy (+0.2\% AMOTA, +0.6\% MOTA) and reduces latency (-3.4ms), verifying its effectiveness in avoiding invalid cost calculations.
        \textcolor{black}{Specifically, it enhances absolute position constraints, preventing wrong matches from redundant detections (-1223 FP).}
        % \textcolor{blue}{This enhancement also reveals that the key to improving tracker performance is to focus on improving the filtering and correlation quality at each small voxel, rather than operating globally.}
        
        \textbf{The effect of the Lightweight Filter.}
        Table \ref{table:nu_ablation} indicates combining the lightweight filter enhances real-time performance (-0.8ms).
        \textcolor{black}{However, the inner limited-step sliding window discards long-term temporal information for time-invariant states, slightly impacting estimation quality and overall tracking accuracy (-0.1\% MOTA).}
        % It is worth mentioning that the lightweight filter also has a slight negative impact on accuracy (-0.1\% MOTA), due to the increase in FP and IDS.
        % \textcolor{blue}{However, the limited-step sliding window is used to filter the temporal time-invariant state, this discards historical information, which to some extent damages the estimation quality. Lightweight Filter also results in a decrease in MOTA (-0.1\%).}

        \textbf{The effect of the Confidence-Count Mixed Lifecycle.} 
        Table \ref{table:nu_ablation} shows our lifecycle module (EXP4) increases processing speed by 4.8 FPS over EXP3, indicating more efficient tracklet management.
        It also improved AMOTA and MOTA by 0.4\% and 0.3\% respectively, validating its efficacy.
        \textcolor{black}{As shown in Table \ref{table:nu_life}, our tracklet termination strategy (line 3) outperforms the original score refinement~\cite{benbarka2021score} (line 2) with a 2.7\% AMOTA increase and 2366 fewer FN, boosting tracker robustness against mismatch scenarios (occlusion, etc.).
        The smoother score prediction (line 4) further enhances tracking performance (+0.4\% AMOTA, +0.1\% MOTA, -926 FN).}
        
        % \textcolor{blue}{
        % Additionally, compared to score refinement, we use a smoother score prediction function, which enhances the estimation quality of trajectories. We also use the average trajectory score to delete trajectories, which enhances the tracker’s robustness to occlusion and mismatch situations. As shown in \textbf{TABLE VII}, these operations help reduce FN.
        % In summary, the key to improving overall accuracy by Confidence-Count Mixed Lifecycle is mainly due to its ability to suppress invalid trajectories early and recall more mismatched trajectories.}

        \textbf{The effect of the Parallelization.}
        \textcolor{black}{Without compromising accuracy or violating online tracking principles}, the multi-processing parallelization effectively alleviates the inherent serial logic of the TBD framework and improves the inference speed by 8.2\% (-3.1ms).
        
        \begin{figure}[t]
        \centering
        %\framebox{\parbox{3in}{We suggest that you use a text box to insert a graphic (which is ideally a 300 dpi TIFF or EPS file, with all fonts embedded) because, in a document, this method is somewhat more stable than directly inserting a picture.}}
        \includegraphics[width=1\linewidth]{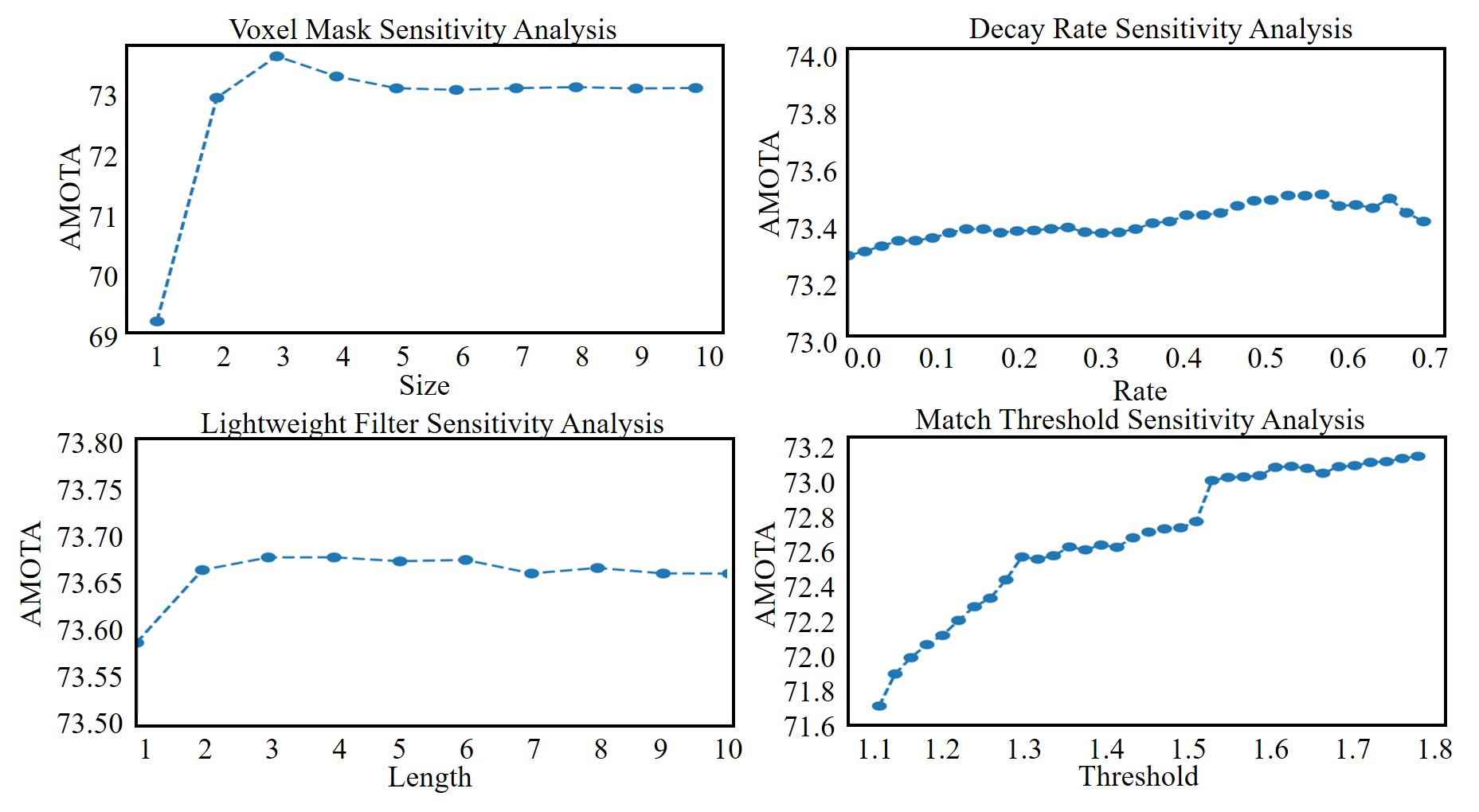}
        \caption{The comparison of the accuracy under distinct newly introduced hyperparameter.
        No category-specific technique is performed, all categories are applied the same parameter. }
        \label{fig:f4}
        \end{figure}

        \subsection{Hyperparameter Sensitivity Analysis}
        \label{Hyperparameter}
        The plethora of hand-crafted hyperparameters poses a fundamental limitation to the TBD framework.
        In response, we conduct a sensitivity analysis on newly introduced hyperparameters \textcolor{black}{on the nuScenes val set}.
        Retaining other parameters, the linear search rule is executed for filter length $l_{lw}$, voxel mask size $\theta_{vm}$, decay rate $\sigma$ and match threshold $\theta_{fm}$.
        A critical finding in Fig. \ref{fig:f4} is that Fast-Poly showcases insensitivity to all parameters except $\theta_{fm}$, with an average fluctuation of less than 0.4\% in AMOTA.
        The pronounced decrease in $\theta_{vm}$ at 1 stems from erroneous matches induced by artificially narrowing the matching space.
        Furthermore, $\theta_{fm}$ emerges as a pivotal factor for all trackers, manifesting sensitivity across categories~\cite{li2023poly, kim2021eagermot, zhang2023bytetrackv2}.
        These findings underscore the resilience of our proposed method in parameter selection.
        The optimal hyperparameters are delineated in \cref{implementation}.

        % \begin{table}[t]
        % \centering
        % \caption{A comparison on various pre-processing modules for on nuScenes val set.
        % The scale factor in Scale-NMS~\cite{huang2021bevdet} is 1.1.
        % \textbf{BEV} means the Bird’s-Eye View representation space.
        % }
        % \label{table:nu_pre}
        % \renewcommand{\arraystretch}{1}
        % \begin{tabular}{ccccc}
        % \toprule
        % \multicolumn{1}{c}{\textbf{Category}} & \textbf{NMS} & \textbf{Metric} & \textbf{AMOTA}$\uparrow$ & \textbf{FP}$\downarrow$ \\ \midrule
        % \multirow{3}{*}{\textit{Bic}} & Standard-NMS                            & $IoU_{bev}$      & 57.1     & 241      \\
        %  & Standard-NMS       & ${A\text{-}gIoU}_{bev}$      & 56.5     & 202    \\
        %   & Scale-NMS~\cite{huang2021bevdet}      & ${A\text{-}gIoU}_{bev}$      & \textbf{57.2}     & \textbf{189}    \\
        % \bottomrule
        % \end{tabular}
        % \end{table}
	\section{Conclusion}
        In this work, we proposed an efficient and powerful polyhedral framework for 3D MOT, termed Fast-Poly.
        To ensure consistency in both accuracy and latency, Fast-Poly integrates three core principles to enhance the baseline Poly-MOT, including:
        (1) Alignment: 
        By aligning objects, our proposed A-GIoU addresses the detrimental effect of 3D rotation on similarity calculation.
        (2) Densification: 
        Through soft lifecycle management, voxel mask, and lightweight filter, we enhance the computational efficiency of trajectory maintenance, cost matrix construction, and filter estimation, respectively.
        (3) Parallelization:
        Based on multi-processing technology, we perform parallel the prediction and pre-processing modules, effectively alleviating the serial defects of the TBD framework.
        We conduct extensive experiments on two large-scale tracking benchmarks.
        Fast-Poly achieves state-of-the-art performance (75.8\% AMOTA and 34.2 FPS) among all methods on nuScenes and demonstrates superior real-time and accuracy performance (63.6\% MOTA and 35.5 FPS) on Waymo.
        Fast-Poly is open source and hopes to contribute to the community.

	%%%%%%%%%%%%%%%%%%%%%%%%%%%%%%%%%%%%%%%%%%%%%%%%%%%%%%%%%%%%%%%%%%%%%%%%%%%%%%%%
	
	\bibliographystyle{IEEEtran}
	\bibliography{IEEEabrv,myref.bib}

\begin{thebibliography}{10}
\providecommand{\url}[1]{#1}
\csname url@samestyle\endcsname
\providecommand{\newblock}{\relax}
\providecommand{\bibinfo}[2]{#2}
\providecommand{\BIBentrySTDinterwordspacing}{\spaceskip=0pt\relax}
\providecommand{\BIBentryALTinterwordstretchfactor}{4}
\providecommand{\BIBentryALTinterwordspacing}{\spaceskip=\fontdimen2\font plus
\BIBentryALTinterwordstretchfactor\fontdimen3\font minus \fontdimen4\font\relax}
\providecommand{\BIBforeignlanguage}[2]{{%
\expandafter\ifx\csname l@#1\endcsname\relax
\typeout{** WARNING: IEEEtran.bst: No hyphenation pattern has been}%
\typeout{** loaded for the language `#1'. Using the pattern for}%
\typeout{** the default language instead.}%
\else
\language=\csname l@#1\endcsname
\fi
#2}}
\providecommand{\BIBdecl}{\relax}
\BIBdecl

\bibitem{li2023poly}
X.~Li, T.~Xie, D.~Liu, J.~Gao, K.~Dai, Z.~Jiang, L.~Zhao, and K.~Wang, ``Poly-mot: A polyhedral framework for 3d multi-object tracking,'' in \emph{IROS}.\hskip 1em plus 0.5em minus 0.4em\relax IEEE, 2023, pp. 9391--9398.

\bibitem{li2023camo}
L.~Wang, X.~Zhang, W.~Qin, X.~Li, J.~Gao, L.~Yang, Z.~Li, J.~Li, L.~Zhu, H.~Wang \emph{et~al.}, ``Camo-mot: Combined appearance-motion optimization for 3d multi-object tracking with camera-lidar fusion,'' \emph{T-ITS}, 2023.

\bibitem{kim2021eagermot}
A.~Kim, A.~O{\v{s}}ep, and L.~Leal-Taix{\'e}, ``Eagermot: 3d multi-object tracking via sensor fusion,'' in \emph{ICRA}.\hskip 1em plus 0.5em minus 0.4em\relax IEEE, 2021, pp. 11\,315--11\,321.

\bibitem{pang2022simpletrack}
Z.~Pang, Z.~Li, and N.~Wang, ``Simpletrack: Understanding and rethinking 3d multi-object tracking,'' in \emph{ECCVW}.\hskip 1em plus 0.5em minus 0.4em\relax Springer, 2022, pp. 680--696.

\bibitem{gwak2022minkowski}
J.~Gwak, S.~Savarese, and J.~Bohg, ``Minkowski tracker: A sparse spatio-temporal r-cnn for joint object detection and tracking,'' \emph{arXiv preprint arXiv:2208.10056}, 2022.

\bibitem{sadjadpour2023shasta}
T.~Sadjadpour, J.~Li, R.~Ambrus, and J.~Bohg, ``Shasta: Modeling shape and spatio-temporal affinities for 3d multi-object tracking,'' \emph{RA-L}, 2023.

\bibitem{zaech2022learnable}
J.-N. Zaech, A.~Liniger, D.~Dai, M.~Danelljan, and L.~Van~Gool, ``Learnable online graph representations for 3d multi-object tracking,'' \emph{RA-L}, vol.~7, no.~2, pp. 5103--5110, 2022.

\bibitem{caesar2020nuscenes}
H.~Caesar, V.~Bankiti, A.~H. Lang, S.~Vora, V.~E. Liong, Q.~Xu, A.~Krishnan, Y.~Pan, G.~Baldan, and O.~Beijbom, ``nuscenes: A multimodal dataset for autonomous driving,'' in \emph{CVPR}, 2020, pp. 11\,621--11\,631.

\bibitem{sun2020scalability}
P.~Sun, H.~Kretzschmar, X.~Dotiwalla, A.~Chouard, V.~Patnaik, P.~Tsui, J.~Guo, Y.~Zhou, Y.~Chai, B.~Caine \emph{et~al.}, ``Scalability in perception for autonomous driving: Waymo open dataset,'' in \emph{CVPR}, 2020, pp. 2446--2454.

\bibitem{weng20203d}
X.~Weng, J.~Wang, D.~Held, and K.~Kitani, ``3d multi-object tracking: A baseline and new evaluation metrics,'' in \emph{IROS}.\hskip 1em plus 0.5em minus 0.4em\relax IEEE, 2020, pp. 10\,359--10\,366.

\bibitem{wang2022deepfusionmot}
X.~Wang, C.~Fu, Z.~Li, Y.~Lai, and J.~He, ``Deepfusionmot: A 3d multi-object tracking framework based on camera-lidar fusion with deep association,'' \emph{RA-L}, vol.~7, no.~3, pp. 8260--8267, 2022.

\bibitem{PC3T}
H.~Wu, W.~Han, C.~Wen, X.~Li, and C.~Wang, ``3d multi-object tracking in point clouds based on prediction confidence-guided data association,'' \emph{T-ITS}, vol.~23, no.~6, pp. 5668--5677, 2021.

\bibitem{liu2023fasttrack}
C.~Liu, H.~Li, and Z.~Wang, ``Fasttrack: A highly efficient and generic gpu-based multi-object tracking method with parallel kalman filter,'' \emph{IJCV}, pp. 1--21, 2023.

\bibitem{benbarka2021score}
N.~Benbarka, J.~Schr{\"o}der, and A.~Zell, ``Score refinement for confidence-based 3d multi-object tracking,'' in \emph{IROS}.\hskip 1em plus 0.5em minus 0.4em\relax IEEE, 2021, pp. 8083--8090.

\bibitem{ding20233dmotformer}
S.~Ding, E.~Rehder, L.~Schneider, M.~Cordts, and J.~Gall, ``3dmotformer: Graph transformer for online 3d multi-object tracking,'' in \emph{ICCV}, 2023, pp. 9784--9794.

\bibitem{zhang2023bytetrackv2}
Y.~Zhang, X.~Wang, X.~Ye, W.~Zhang, J.~Lu, X.~Tan, E.~Ding, P.~Sun, and J.~Wang, ``Bytetrackv2: 2d and 3d multi-object tracking by associating every detection box,'' 2023.

\bibitem{mutr3d}
T.~Zhang, X.~Chen, Y.~Wang, Y.~Wang, and H.~Zhao, ``Mutr3d: A multi-camera tracking framework via 3d-to-2d queries,'' in \emph{CVPR}, 2022, pp. 4537--4546.

\bibitem{pftrack}
Z.~Pang, J.~Li, P.~Tokmakov, D.~Chen, S.~Zagoruyko, and Y.-X. Wang, ``Standing between past and future: Spatio-temporal modeling for multi-camera 3d multi-object tracking,'' in \emph{CVPR}, 2023, pp. 17\,928--17\,938.

\bibitem{yin2021center}
T.~Yin, X.~Zhou, and P.~Krahenbuhl, ``Center-based 3d object detection and tracking,'' in \emph{CVPR}, 2021, pp. 11\,784--11\,793.

\bibitem{zhou2020tracking}
X.~Zhou, V.~Koltun, and P.~Kr{\"a}henb{\"u}hl, ``Tracking objects as points,'' in \emph{ECCV}.\hskip 1em plus 0.5em minus 0.4em\relax Springer, 2020, pp. 474--490.

\bibitem{bernardin2006multiple}
K.~Bernardin, A.~Elbs, and R.~Stiefelhagen, ``Multiple object tracking performance metrics and evaluation in a smart room environment,'' in \emph{Sixth IEEE International Workshop on Visual Surveillance, in conjunction with ECCV}, vol.~90, no.~91.\hskip 1em plus 0.5em minus 0.4em\relax Citeseer, 2006.

\bibitem{luiten2021hota}
J.~Luiten, A.~Osep, P.~Dendorfer, P.~Torr, A.~Geiger, L.~Leal-Taix{\'e}, and B.~Leibe, ``Hota: A higher order metric for evaluating multi-object tracking,'' \emph{IJCV}, vol. 129, pp. 548--578, 2021.

\bibitem{huang2021bevdet}
J.~Huang, G.~Huang, Z.~Zhu, Y.~Ye, and D.~Du, ``Bevdet: High-performance multi-camera 3d object detection in bird-eye-view,'' \emph{arXiv preprint arXiv:2112.11790}, 2021.

\bibitem{liu2023bevfusion}
Z.~Liu, H.~Tang, A.~Amini, X.~Yang, H.~Mao, D.~L. Rus, and S.~Han, ``Bevfusion: Multi-task multi-sensor fusion with unified bird's-eye view representation,'' in \emph{ICRA}.\hskip 1em plus 0.5em minus 0.4em\relax IEEE, 2023, pp. 2774--2781.

\bibitem{chen2022scaling}
Y.~Chen, J.~Liu, X.~Qi, X.~Zhang, J.~Sun, and J.~Jia, ``Scaling up kernels in 3d cnns,'' \emph{arXiv preprint arXiv:2206.10555}, 2022.

\bibitem{wu2022casa}
H.~Wu, J.~Deng, C.~Wen, X.~Li, C.~Wang, and J.~Li, ``Casa: A cascade attention network for 3-d object detection from lidar point clouds,'' \emph{T-GRS}, vol.~60, pp. 1--11, 2022.

\bibitem{shi2020pv}
S.~Shi, C.~Guo, L.~Jiang, Z.~Wang, J.~Shi, X.~Wang, and H.~Li, ``Pv-rcnn: Point-voxel feature set abstraction for 3d object detection,'' in \emph{CVPR}, 2020, pp. 10\,529--10\,538.

\bibitem{kuhn1955hungarian}
H.~W. Kuhn, ``The hungarian method for the assignment problem,'' \emph{Naval research logistics quarterly}, vol.~2, no. 1-2, pp. 83--97, 1955.

\bibitem{sutherland1974reentrant}
I.~E. Sutherland and G.~W. Hodgman, ``Reentrant polygon clipping,'' \emph{Communications of the ACM}, vol.~17, no.~1, pp. 32--42, 1974.

\bibitem{graham1972efficient}
R.~L. Graham, ``An efficient algorithm for determining the convex hull of a finite planar set,'' \emph{Info. Proc. Lett.}, vol.~1, pp. 132--133, 1972.

\bibitem{rezatofighi2019generalized}
H.~Rezatofighi, N.~Tsoi, J.~Gwak, A.~Sadeghian, I.~Reid, and S.~Savarese, ``Generalized intersection over union: A metric and a loss for bounding box regression,'' in \emph{CVPR}, 2019, pp. 658--666.

\bibitem{bai2022faster}
C.~Bai, T.~Xiao, Y.~Chen, H.~Wang, F.~Zhang, and X.~Gao, ``Faster-lio: Lightweight tightly coupled lidar-inertial odometry using parallel sparse incremental voxels,'' \emph{RA-L}, vol.~7, no.~2, pp. 4861--4868, 2022.

\bibitem{van2011numpy}
S.~Van Der~Walt, S.~C. Colbert, and G.~Varoquaux, ``The numpy array: a structure for efficient numerical computation,'' \emph{Computing in science \& engineering}, vol.~13, no.~2, pp. 22--30, 2011.

\bibitem{cai2018cascade}
Z.~Cai and N.~Vasconcelos, ``Cascade r-cnn: Delving into high quality object detection,'' in \emph{CVPR}, 2018, pp. 6154--6162.

\bibitem{bai2022transfusion}
X.~Bai, Z.~Hu, X.~Zhu, Q.~Huang, Y.~Chen, H.~Fu, and C.-L. Tai, ``Transfusion: Robust lidar-camera fusion for 3d object detection with transformers,'' in \emph{CVPR}, 2022, pp. 1090--1099.

\bibitem{wu20213d}
H.~Wu, W.~Han, C.~Wen, X.~Li, and C.~Wang, ``3d multi-object tracking in point clouds based on prediction confidence-guided data association,'' \emph{T-ITS}, vol.~23, no.~6, pp. 5668--5677, 2021.

\bibitem{DETR3D}
Y.~Wang, V.~C. Guizilini, T.~Zhang, Y.~Wang, H.~Zhao, and J.~Solomon, ``Detr3d: 3d object detection from multi-view images via 3d-to-2d queries,'' in \emph{CoRL}.\hskip 1em plus 0.5em minus 0.4em\relax PMLR, 2022, pp. 180--191.

\bibitem{cc3dt}
T.~Fischer, Y.-H. Yang, S.~Kumar, M.~Sun, and F.~Yu, ``Cc-3dt: Panoramic 3d object tracking via cross-camera fusion,'' 2022.

\end{thebibliography}
	
\end{document}